\documentclass[letterpaper]{article} % DO NOT CHANGE THIS
\usepackage[submission]{aaai25}  % DO NOT CHANGE THIS
\usepackage{times}  % DO NOT CHANGE THIS
\usepackage{helvet}  % DO NOT CHANGE THIS
\usepackage{courier}  % DO NOT CHANGE THIS
\usepackage[hyphens]{url}  % DO NOT CHANGE THIS
\usepackage{graphicx} % DO NOT CHANGE THIS
\usepackage{natbib}  % DO NOT CHANGE THIS AND DO NOT ADD ANY OPTIONS TO IT
\usepackage{caption} % DO NOT CHANGE THIS AND DO NOT ADD ANY OPTIONS TO IT
\usepackage{algorithm}
\usepackage{algorithmic}

\usepackage{amssymb}  % HAVE ADDED THIS!!!!!!!!!
\usepackage{amsmath}  % HAVE ADDED THIS!!!!!!!!!
\usepackage{xcolor}  % HAVE ADDED THIS!!!!!!!!!
\usepackage{comment}  % HAVE ADDED THIS!!!!!!!!!

\usepackage{pgfplots}    %HAVE ADDED THIS!!!!!!!!!!!!
\usepgflibrary{plotmarks}    %HAVE ADDED THIS!!!!!!!!!!!!
\usepackage{subcaption}  %HAVE ADDED THIS!!!!!!!!!!!!!!

\DeclareMathOperator{\E}{\mathbb{E}}  % HAVE ADDED THIS!!!!!!!!!
\DeclareMathOperator{\IndicatorFunc}{\mathbf{1}}  % HAVE ADDED THIS!!!!!!!!!

\usepackage{newfloat}
\usepackage{listings}
\DeclareCaptionStyle{ruled}{labelfont=normalfont,labelsep=colon,strut=off} % DO NOT CHANGE THIS
\floatstyle{ruled}
\newfloat{listing}{tb}{lst}{}
\floatname{listing}{Listing}
\title{Markov Balance Satisfaction Improves Performance\\ in Strictly Batch Offline Imitation Learning}
\author{
    Rishabh Agrawal\textsuperscript{\rm 1},
    Nathan Dahlin\textsuperscript{\rm 2},
    Rahul Jain\textsuperscript{\rm 1},
    Ashutosh Nayyar\textsuperscript{\rm 1}
}
\affiliations{
    \textsuperscript{\rm 1} University of Southern California\\
    \textsuperscript{\rm 2} University at Albany, SUNY\\
    rishabha@usc.edu, ndahlin@albany.edu, \{rahul.jain, ashutosn\}@usc.edu
}

\usepackage{bibentry}
\makeatletter
\newenvironment{customlegend}[1][]{%
    \begingroup
    \let\addlegendimage=\pgfplots@addlegendimage
    \let\addlegendentry=\pgfplots@addlegendentry
    \pgfplots@init@cleared@structures
    \pgfplotsset{#1}%
}{%
    \pgfplots@createlegend
    \endgroup
}%

\definecolor{bcColor}{RGB}{0, 255, 0}
\definecolor{vdiceColor}{RGB}{255, 0, 0}
\definecolor{rcalColor}{RGB}{110,117,14}
\definecolor{edmColor}{RGB}{0,255,255}
\definecolor{avrilColor}{RGB}{255,0,255}
\definecolor{dsfnColor}{RGB}{0, 0, 255}
\definecolor{ckilColor}{RGB}{94,60,153}
\definecolor{iqlearnColor}{RGB}{255, 255, 0}
\definecolor{expertColor}{RGB}{240,100,10}
\definecolor{randomColor}{RGB}{0,0,0}
\definecolor{softdiceColor}{RGB}{0, 0, 255}
\definecolor{odiceColor}{RGB}{0,255,255}
\makeatother

\pgfplotsset{compat=1.18}

\newtheorem{theorem}{Theorem}

\begin{document}

\maketitle

\begin{abstract}
Imitation learning (IL) is notably effective for robotic tasks where directly programming behaviors or defining optimal control costs is challenging. In this work, we address a scenario where the imitator relies solely on observed behavior and cannot make environmental interactions during learning. It does not have additional supplementary datasets beyond the expert's dataset nor any information about the transition dynamics. Unlike state-of-the-art (SOTA) IL methods, this approach tackles the limitations of conventional IL by operating in a more constrained and realistic setting. Our method uses the Markov balance equation and introduces a novel conditional density estimation-based imitation learning framework. It employs conditional normalizing flows for transition dynamics estimation and aims at satisfying a balance equation for the environment. 
%\rishabh{To add a line on convergence of CNF}. 
Through a series of numerical experiments on Classic Control and MuJoCo environments, we demonstrate consistently superior empirical performance compared to many SOTA IL algorithms.
\end{abstract}

\section{Introduction}\label{sec:intro}

Reinforcement Learning (RL) has provided us with some very notable achievements over the past decade from mastering complex games \citep{mnih2015human,silver2016mastering,vinyals2019grandmaster} to advancing protein structure prediction systems \citep{jumper2021highly}, and now proficiency at coding and high school-level mathematics. And yet, all of these achievements are built on a fundamental hypothesis that the ``reward is enough" \citep{silver2021reward}. But in the real world, when only data is available, there is no natural reward model available. Countless hours are spent on reward engineering. In fact, lately, there has been tremendous progress in building reward models for language and multi-modal generative AI systems. And yet, they have led to their own set of challenges, namely reward hacking and over-optimization, lack of diversity and robustness in learned policies. 

Classically, in imitation learning, this was sought to be addressed via inverse RL methods \citep{AndrewNG1, AndrewNG2}, e.g., the MaxEntropy-IRL algorithm \citep{maxEntRL} that first estimates a reward function from demonstration data and then uses it with RL algorithms to find near-optimal policies. Unfortunately, this has two limitations: First, the reward function estimation problem is ill-posed \citep{baheri2023understanding} and any additional criterion introduce estimation errors. Second, the demonstrator (e.g., a human subject) may not be optimizing with respect to any reward function at all! Additionally, IRL algorithms are computationally intensive, particularly in high-dimensional state spaces, limiting their scalability \citep{barnes2024massively}. 
%Bayesian IRL methods also require substantial prior information, and inaccuracies in this prior knowledge can impact learning outcomes \citep{adams2022survey}. Addressing these limitations is crucial for advancing IRL's practical applicability.
Thus, there is a need to develop Imitation Learning (IL) algorithms that don't depend on reward estimation as a first step.

Behavioral Cloning (BC) is one such classical IL method  \citep{firstBC}. It is inspired by supervised learning and learns a map from states to actions from trajectory data. Unfortunately, it does not account for the fact that such trajectory data often satisfies a Markov balance equation (MBE). In fact, this MBE is the only mathematical structure that ties the trajectory data together. And if we don't use it, we can expect we will not do as well as possible. Thus, it is not unexpected that the BC method is known to be vulnerable to error propagation and covariate shift issues, thus constraining its generalization capabilities \citep{bcLimitation1}. This is typically sought to be addressed by allowing for additional online interactions  with the environment or the demonstrator, or using supplementary data  \citep{BCMod1, BCMod2}. Unfortunately, in many practical situations, this is simply not a possibility such as in autonomous vehicles and healthcare, where engaging directly with the environment is often infeasible due to safety concerns or high costs. Therefore, we must learn from the offline data we already have. A premise of ours is that just like in training of LLMs, a combination of supervised fine-tuning and RL with human feedback improves LLM performance, in the same way, an imitation learning method that combines behavior cloning policy with accounting for Markovian dynamics will perform better as well.

At the core of imitation learning is distribution matching, which treats state-action pairs from expert demonstrations as samples drawn from a target distribution. The objective is to develop a policy that minimizes the divergence between this target distribution and the distribution induced by the imitator's policy. Adversarial Imitation Learning (AIL) methods use this principle \citep{gail, fu2018AIL, fDivAIL}, but they face a significant limitation: estimating density ratios requires on-policy samples from the environment. This need for continuous interaction with the environment makes AIL methods impractical when only offline data is available.

To address the availability of limited expert data, \citep{yue2023clare, xu2022discriminator} propose a supplementary data framework in imitation learning. This method uses additional data, cheaply gathered by executing suboptimal policies, to augment the expert dataset. However, such supplementary data may not be available, and in fact may even cause distribution shifts due to out-of-expert-distribution trajectories, which may degrade model performance. 

In this paper, we focus on the strictly batch imitation learning problem (i.e., no further interaction or supplementary data) and propose a novel approach to imitation learning that does not do \textit{reward model estimation}, or \textit{distribution matching} between occupation measures via on-policy samples. Instead, it learns a policy that minimizes a loss function related to satisfaction of the Markov balance equation, which is the one and only fundamental relationship we know about the trajectory data. This Markov balance equation involves two conditional state-action transition density functions which need to be learnt from data. Density estimation, and especially conditional density estimation is rather tricky, especially for continuous state and action space settings. Fortunately, recently developed normalizing flows methods \citep{dinh2017density} have proven remarkably well-suited for this problem. Furthermore, we regularize the MBE loss around a behavioral cloning policy, thus combining supervised and RL methods for imitation learning in a natural manner. This \textit{novel} combination enables excellent numerical performance compared to state-of-the-art IL/IRL/AIL methods in the considered settings across a range of Classic Control and MuJoCo tasks.

\begin{figure*} % Use figure* for spanning both columns
     \centering
     \includegraphics[width=1\linewidth]{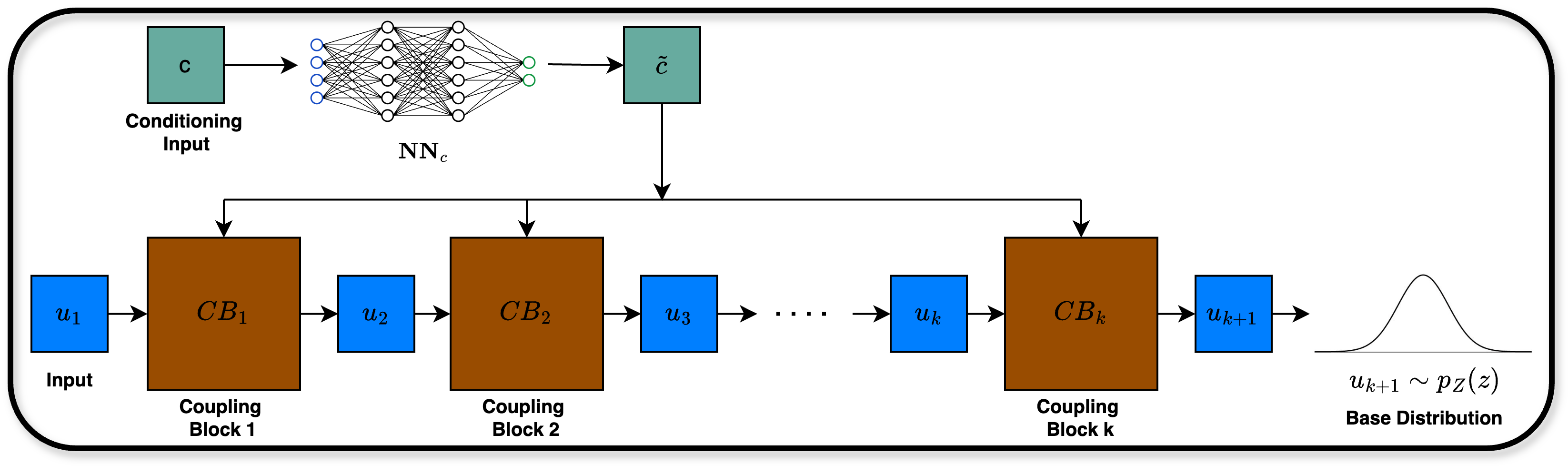}
     \caption{A high-level description of Conditional Normalizing Flow, where $CB_{i}$ is the $i^{th}$ coupling block. }%\rjain{REMOVE}}
 \label{fig:conditional_normalizing_flows}
\end{figure*}

%\subsubsection*{Related Work.}
%\parskip
\noindent\textbf{Related Work.} \textbf{\em Offline IRL.} To circumvent costly online environmental interactions in classic IRL, offline IRL aims to infer a reward function and recover the expert policy solely from a static dataset without accessing the environment. \citep{klein2011} introduced \textit{LSTD-$\mu$}, extending classic apprenticeship learning \citep{AndrewNG1} to batch and off-policy cases for computing feature expectations, while \citep{klein2012} developed a score function-based classification algorithm to output the reward function.
\citep{lee2019} propose \textit{DSFN}, using a transition-regularized imitation network for an initial policy close to expert behavior. However, these methods assume complete knowledge of reward features, which is often impractical for complex problems due to the problem-dependent nature of feature selection \citep{aroraAndDoshi2021}. \textit{RCAL} \citep{bcLimitation2} employs a boosting method to minimize a large margin objective with a regularization term, avoiding feature selection steps. \citep{chan2021} introduced \textit{AVRIL}, jointly learning an approximate posterior distribution over reward and policy. \citep{garg2021iq} proposed \textit{IQ-Learn}, using a learned soft Q-function to represent both reward and policy implicitly. Despite these advances, they struggle with covariate shift and reward extrapolation errors, impacting performance in novel environments \citep{clare}. \textit{CLARE} \citep{clare} incorporates conservatism in reward estimation but performs poorly with low-quality transition samples \citep{Siliang} and requires additional diverse datasets unavailable in our setting. %\rjain{Too long: shorten and save space.}

\noindent\textbf{\em Offline IL.} The EDM approach \citep{SBIL} captures the expert's state occupancy measure by training an explicit energy-based model but faces significant limitations and is unsuitable for continuous action domains \citep{sbilCritique}. DICE (DIstribution Correction Estimation) methods perform stationary distribution matching of state-action pairs between the learner's policy and expert demonstrations. \citep{VDICE_iclr} introduces \textit{ValueDICE}, which minimizes the Donsker-Varadhan representation of KL divergence between these stationary distributions but suffers from biased gradient estimates due to logarithmic terms applied to expectations. SoftDICE \citep{sun2021softdice} addresses these limitations by using the Earth-Mover Distance (EMD) for distribution matching, eliminating problematic logarithmic and exponential terms. DemoDICE \citep{kim2022demodice} and SMODICE \citep{ma2022versatile} assume additional demonstration data of unknown degrees of optimality beyond expert's data to deal with the narrow support of the expert data distribution, but such information is unavailable in our setup. DICE methods involve two gradient terms for value function learning: the forward gradient (on the current state) and the backward gradient (on the next state). However, conflicting directions between these gradients can lead to performance degradation \citep{mao2024odice}. To address this, \citep{mao2024odice} introduces ODICE, an orthogonal-gradient update method that projects the backward gradient onto the normal plane of the forward gradient, ensuring effective state-action-level constraints and better convergence. 
%2/3 \rjain{Shorten this also if you can, and save space.}

Our work is clearly differentiated from prior work in not doing stationary distribution matching, but instead in first using normalizing flows in conditional state-action density estimation, and then using these in ensuring Markov balance equation satisfaction.

\section{Preliminaries}\label{sec:prelim}

\textbf{The Imitation Learning Problem.}
An infinite horizon discounted Markov decision process (MDP) $M$ is defined by the tuple $(S,A,T,r,\gamma)$ with states $s\in S$, actions $a\in A$ and successor states $s' \in S$ drawn from the transition function $T(s'\vert s, a)$. The reward function $r : S \times A \rightarrow \mathbb{R}$ maps state-action pairs to scalar rewards, and $\gamma$ is the discount factor. Policy $\pi$ is a probability distribution over actions conditioned on state and is given by $\pi(a|s) = P_{\pi}(a_{t} = a |  s_{t}= s)$, where $a_{t}\in A$, $s_{t}\in S$, $\forall t = 0, 1, 2, \cdots$. The induced occupancy measure of a policy is given as $\rho_{\pi}(s, a) := \E_{\pi}[\sum_{t=0}^{\infty} \gamma^{t} \IndicatorFunc_{s_{t} = s, a_{t} = a}]$, where the expectation is taken over $a_{t} \sim \pi(\cdot|s_{t})$, $s_{t+1} \sim T(\cdot|s_{t}, a_{t})$ for all $t$, and the initial state $s_0$. The corresponding state-only occupancy measure is given as $\rho_{\pi}(s) = \sum_{a} \rho_{\pi}(s,a)$.
In the \textcolor{black}{offline} imitation learning (IL) framework, the agent is provided with trajectories generated by a demonstration policy $\pi_D$, collected as $D=\{(s_0,a_0), (s_1,a_1), (s_2,a_2),...\}$\textcolor{black}{; and is not allowed any further interaction with the environment.} The data $D$ does \emph{not} include any reward $r_t$ at each time step. Indeed, rather than long-term reward maximization, the IL objective is to learn a policy $\pi^{\star}$ that is close to $\pi_D$ in the following sense \citep{ImitationLearningPresentation}:
\begin{equation}
    \pi^{\star}\in \underset{\pi \in \Pi}{\arg\min}\,\mathbb{E}_{s \sim \rho_\pi} [\mathcal{L}(\pi(\cdot|s), \pi_D(\cdot|s))],
    \label{eq:ILObjective}
\end{equation}
where $\Pi$ is the set of all randomized (Markovian) stationary policies, and $\mathcal{L}$ is a chosen loss function. In practice, \eqref{eq:ILObjective} can only be solved approximately since  $\pi_D$ is unknown and only transitions are observed in the  dataset $D$.

\begin{figure*} % Use figure* for spanning both columns
     \centering
     \includegraphics[width=1\linewidth]{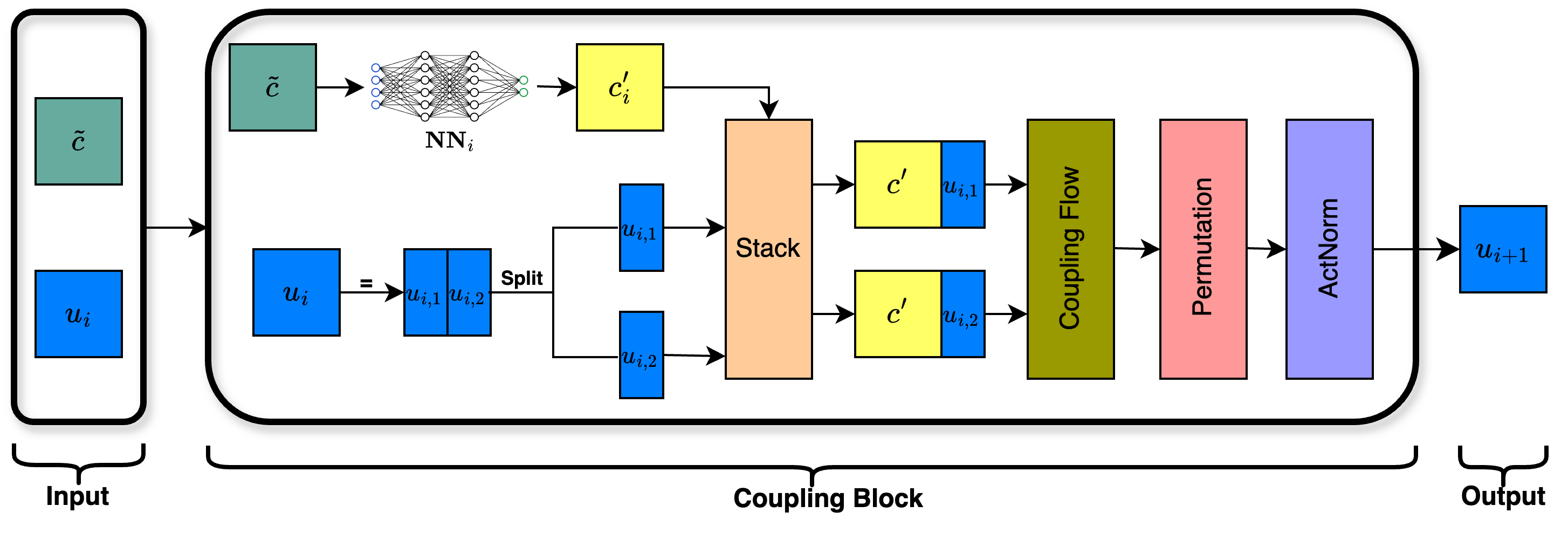}
     \caption{A high-level description of coupling block $CB_{i}$ in conditional density estimation setup.} %\rjain{REMOVE}}
 \label{fig:coupling_block}
\end{figure*}

\textbf{Normalizing Flows.} 
% Thus, we adopt kernel density estimation (KDE), a nonparametric framework for the estimation of general continuous distributions 
% We present an imitation learning framework that centers on the challenging task of conditional density estimation. Despite advancements in statistical theory, this task remains difficult due to uncertainties surrounding suitable parametric families of density functions. Normalizing flows (NFs) emerge as a potent solution, belonging to a class of generative models adept at handling complex probability distributions.
The imitation learning approach we introduce depends on transition density estimation. Despite advancements in statistical theory, conditional density estimation is a difficult problem due to a lack of clarity on what parametric families of density functions are good candidates. We make use of normalizing flows (NFs) to address our density estimation problem. NFs belong to a class of generative models adept at modeling complex probability distributions.
NFs employ a sequence of invertible and differentiable transformations $g_{\phi, k} = g_k \circ g_{k-1} \circ \cdots \circ g_1$, applied to a base distribution $p_Z(z)$, typically a simple distribution, e.g., standard Gaussian. These transformations progressively refine the distribution to accurately model the target data distribution $p_X(x)$ via the mapping $z = g_{\phi, k}(x)$. Consequently, the log-likelihood of the target distribution is formulated as:
\begin{equation}
\log p_X(x) = \log p_Z(z) + \sum_{i=1}^{k} \log \left| \det \frac{\partial g_i}{\partial g_{i-1}} \right|,
\label{eq:NFObjective}
\end{equation}
which facilitates the training of parameters $\phi$ via maximum likelihood. Once the flow is trained, density estimation can be performed by evaluating the right-hand side of \eqref{eq:NFObjective}.

Early contributions such as NICE \citep{dinh2014nice} streamline computations using non-linear additive coupling layers. Building on this, RealNVP \citep{dinh2017density} utilizes affine coupling layers to achieve efficient and precise density estimation and sampling, offering improved expressiveness and flexibility compared to NICE. Masked Autoregressive Flow (MAF) \citep{papamakarios2017masked} replaces these affine coupling layers with autoregressive ones, increasing expressiveness at the expense of single-pass sampling. More recently, Glow \citep{kingma2018glow} advances these concepts further by incorporating invertible 1x1 convolutions, offering greater flexibility within multi-scale architectures.

\section{Markov Balance-based Imitation Learning}\label{sec:framework}

We next describe MBIL, our imitation learning algorithm. A key premise of our algorithm is that  the demonstration trajectory data satisfies a balance equation involving the demonstration policy, the Markov decision process (MDP) transition density, and the induced Markov chain (MC).
We can use this balance equation to guide the agent's learning. Using the balance equation requires estimating certain transition (conditional probability) density functions, which we obtain via the conditional normalizing flow method. 
% Then, the agent's learning problem reduces to identifying policies that minimize 
We describe our approach in Algorithm \ref{alg:MBIL} and present numerical evidence of its efficacy on several benchmark Classic Control and MuJoCo tasks.

\begin{comment}
\begin{figure*}
\begin{subfigure}[b]{0.48\textwidth}
        \includegraphics[width=\linewidth]
        {Plots/Normalizing_Flow_High_Level_Details.png}
        %{Plots/Hopper-v2_high_resolution.png}
        %{AnonymousSubmission/LaTeX/AcrobotPlot.png} % Replace with your figure
        \caption{A high-level description of Conditional Normalizing Flow, where $CB_{i}$ is the $i^{th}$ coupling block.}
        \label{subfig:CNF}
    \end{subfigure}
    \hfill
    \begin{subfigure}[b]{0.48\textwidth}
        \includegraphics[width=\linewidth]
        {Plots/Coupling_Block_Details.png}
        %{Plots/Ant-v2_high_resolution.png}
        %{AnonymousSubmission/LaTeX/CartPolePlot.png} % Replace with your figure
        \caption{A high-level description of coupling block $CB_{i}$ in conditional density estimation setup.}
        \label{subfig:CBi}
    \end{subfigure}
    \hfill
    \caption{Representations of Conditional Normalizing Flow and its components.}
    \label{fig:conditional_normalizing_flow}
\end{figure*}
\end{comment}

\textbf{The Markov Balance Equation.} 
Consider a demonstration policy $\pi_D$ that is used to take actions starting from an initial state $s_0$. Let $T(s'|s,a)$ denote the transition density function of the MDP. Note that $\pi_D$ is a randomized stationary Markovian  policy and $\pi_D(\cdot|s)$ is the   probability distribution of actions at state $s$. Using  policy $\pi_D$ on the underlying MDP induces a Markov chain on the state space $S$ whose transition density is denoted by $P(s'|s)$. This transition density satisfies the following equation which we refer to as the Markov balance equation:
\(
P(s'|s) = \sum_a \pi_D(a|s)T(s'|s,a).
\)
% \textcolor{red}{why sums and not integrals?}
%Unfortunately, this involves a sum, and hence is difficult to use \textcolor{red}{in continuous action domain where sum changes to integral over actions and we only have one action sample from a given state}. 
\textcolor{black}{Unfortunately, this approach involves a summation, which becomes problematic in continuous action domains, where the sum translates to an integral over actions}. 
% and we only have a single action sample for a given state in the dataset.
Therefore, we use the following alternative balance equation which involves the  transition density of the induced Markov chain on the \emph{state-action} space,
\begin{equation}
    P_{\pi_D}(s', a'|s,a) = \pi_D(a'|s')T(s'|s,a).
    % P(s', a'|s,a) = \pi_D(a'|s')T(s'|s,a).
    \label{eqn:policy_balance}
\end{equation}

The above balance equation is the basis of our IL approach. If we can estimate $P_{\pi_D}$ and $T$ in \eqref{eqn:policy_balance} (estimates denoted by $\hat{P}$ and $\hat{T}$ respectively), we can use the balance equation to guide agent's learning. In our approach, we consider a combination of a  \emph{policy-based} loss function that measures the discrepancy between the demonstrator's and learner's policies and a \emph{dynamics-based} loss function that quantifies the discrepancy between the two sides of the balance equation under the learner's policy.

% then infer a policy $\pi_D$ that satisfies it. \textcolor{red}{Unfortunately, the problem is ill-conditioned, and we will need to impose additional criterion such as a regularization term. \rishabh{Need to frame this red part in a way so that there is a link between poicy based and dynamics based loss that follows}}

We consider a class of policies parametrized by $\theta$  and formulate the following optimization problem: 

\begin{comment}
\begin{equation}
%\small
    \begin{gathered}
        \min_{\theta \in \Theta} \E_{s, a, s', a' \sim \rho_{\pi_D}}[\hspace{.05cm}\text{BALANCE}(s', a', s, a)\hspace{.05cm}]\\
        %\int_{(s',a')} \int_{(s,a)} \big[\hat{P}(s',a'|s,a) - \pi_{\theta}(a'|s')\hat{T}(s'|s,a)\big]^{2}
        %\\ \,d\mu(s,a) \,d\mu(s',a') 
        %- \lambda\int_{s'} H(\pi_{\theta}(\cdot|s')) \,d\nu(s').
    \end{gathered}
    \label{eqn:markovBalanceObjective}
\end{equation}
\end{comment}
\begin{equation}
\begin{gathered}
%\hspace{-1cm}
%\min_{\theta \in \Theta} J_{\mathrm{MBIL}}(\pi_{\theta}):=\alpha \underbrace{\mathbb{E}_{s, a \sim \rho_{\pi_{D}}}\left[L\left(\pi_{D}(\cdot \mid s), \pi_{\theta}(\cdot \mid s)\right)\right]}_{\text{Policy Loss}}\\
%+\beta \underbrace{\E_{s, a, s', a' \sim \rho_{\pi_D}}[\text{BALANCE}(s', a', s, a)]}_{\text{Dynamics Loss}}
\min_{\theta \in \Theta} J_{\mathrm{MBIL}}(\pi_{\theta}):= \alpha \underbrace{\E_{s, a, s', a' \sim D}[\text{BALANCE}(s', a', s, a)]}_{\text{Dynamics Loss}}\\
+
\beta \underbrace{\mathbb{E}_{s, a \sim D}\left[\mathcal{L}\left(\pi_{D}(\cdot \mid s), \pi_{\theta}(\cdot \mid s)\right)\right]}_{\text{Policy Loss}}\\
\end{gathered}
\label{eqn:markovBalanceObjective}
\end{equation}
\begin{comment}
\begin{equation*}
\begin{gathered}
\hspace{-10cm}J_{\mathrm{alg}}(\pi)=\\\underbrace{\mathbb{E}_{s' \sim d_{\pi^{\star}}}\left[l_1\left(\pi^{\star}(\cdot \mid s'), \pi_{\theta}(\cdot \mid s')\right)\right]}_{\text{Policy Loss}}\\
+\lambda \underbrace{\mathbb{E}_{d_{s, a, s', a' \sim d_{\pi^{\star}}}}\left[l_2\left(\hat{P}(s',a'|s,a), \pi_{\theta}(a'|s') \hat{T}(s'|s,a)\right)\right]}_{\text{Dynamics Loss}}
\end{gathered}
\end{equation*}
\end{comment}
%, where $l_1$ is a loss function designed to align the agent's policy with the expert's policy and $BALANCE$ is a loss function to enforce the Markov Balance constraint.

% We consider a general framework for imitation learning in which any loss function from the IL literature can be used for $L(\cdot)$. Additionally, we introduce a \textit{novel} dynamics loss term to ensure Markov Balance satisfaction within the given environment.
%\textcolor{red}{Q: instead of writing $\mathbb{E}_{s, a \sim \rho_{\pi_{D}}}$ in the objective above, why not just write $\mathbb{E}_{s, a \sim D}$. Similarlry, in the balance term use $\mathbb{E}_{s, a,s',a' \sim D}$? }
%\rishabh{I thought having $\rho_{\pi_{D}}$ ensures the true loss and in practical scenarios then like in equation (5), we can approximate that by samples from D}

\textcolor{black}{Examples of policy loss function $\mathcal{L}(\cdot)$ include KL divergence between the policies (which amounts to maximizing the log-likelihood of the expert's actions in the states observed in the demonstration data)} or  the mean-squared error between the agent's and the expert's actions, among other possibilities. The $BALANCE$  function introduces a dynamics-based loss term designed to enforce the Markov Balance constraint and is defined as follows:
%In our setup, we consider $l_1 = -\text{log}\pi_{\theta}(\cdot \mid s')$ and $l_2 = \left[log\hat{P}((s',a')|(s,a)) - log\pi_{\theta}(a'|s') - log\hat{T}(s'|(s,a))\right]^{2}$.
\begin{equation*}
%\small
    \begin{gathered}
        \text{BALANCE}(s', a', s, a) = \\
        \big[\log\hat{P}(s',a'|s,a) -  \log \pi_{\theta}(a'|s') - \log\hat{T}(s'|s,a)\big]^{2}.
        %\\ \,d\mu(s,a) \,d\mu(s',a') 
        %- \lambda\int_{s'} H(\pi_{\theta}(\cdot|s')) \,d\nu(s').
    \end{gathered}
\end{equation*}
% and $s, a, s', a' \sim \rho_{\pi_D}$ is equivalent to $a \sim \pi_{D}(\cdot|s)$, $s' \sim T(\cdot|s, a)$, and $a' \sim \pi_{D}(\cdot|s').$
%In \eqref{eqn:markovBalanceObjective}, the squared loss term ensures that the balance equation is satisfied approximately. It is a simple but novel loss function rarely used in imitation learning in conjunction with a balance equation. 
$\Theta$ is a given parameter set. The parameters could be weights of a neural network, for example. $\alpha$ and $\beta$ are parameters that represent the relative priorities of the dynamics-based term and the policy-based term, respectively.

\textbf{Remarks. }%\rishabh{Please review this paragraph, I've added some more points} 
In Offline IL-compatible DICE methods like ValueDICE, the derivation of the  off-policy objective assumes that the Markov chain induced by the learned policy is always ergodic and that the state-action distribution is stationary. However, these assumptions often fail during training due to early terminations from environment resets triggered by adverse states or fatal actions leading to termination bias \cite{sun2021softdice}. Algorithms such as IQ-Learn and SoftDICE address this by adding an indicator for absorbing states and modifying the Bellman operator, but this can introduce reward bias, as the value of absorbing states should be learned rather than assumed to be zero \cite{kostrikov2018discriminator}. LS-IQ \cite{al-hafez2023lsiq} improves upon IQ-Learn by correcting reward bias with a modified inverse Bellman operator but performs poorly in pure offline settings, especially on MuJoCo tasks. 
%In contrast, our approach avoids the off-policy derivation framework and the associated termination and reward bias. It uses all transitions in the training data to compute the Markov balance-based loss. 
In contrast, our method avoids the off-policy derivation framework and employs the Markov Balance Equation instead. The rationale is straightforward: \textit{every} transition captured in the dataset, including those leading to the terminal state, must adhere to this equation. As a result, we eliminate the biases that stationary distribution matching algorithms often face.
This also eliminates the need for a discount factor $\gamma$ (typically required in DICE methods and may influence their performance), and simplifies training by avoiding a complex alternating max-min optimization (typical in DICE methods) with a single optimization procedure.

\textbf{Transition Density Estimation}
%We now address the density (mass in discrete case) estimation methodology for the two conditional densities $P_{\pi_{D}}$ and $T$. We will start with the discrete setting, where the estimation is intuitive, and then address the continuous setting.
We now introduce methodology for estimation of the the two conditional densities $P_{\pi_{D}}$ and $T$. 
%We will start with the discrete setting, where the estimation is intuitive, and then address the continuous setting.
%\rishabh{To remove Discrete Spaces counting measure part}
\begin{comment}
\textbf{Discrete Spaces.}
When both the state and action spaces are discrete, the estimates $\hat{P}$ and $\hat{T}$ can be calculated as:

\begin{equation}
    \begin{gathered}
    \hat{T}(s'|s,a):= \frac{\eta (s, a,s' )}{\eta (s, a)},~~\text{and}~~
    \\
    \hat{P}(s', a'|s, a):= \frac{\eta (s, a, s', a')}{\eta (s, a)}
    \end{gathered}
    \label{eqn:discreteProbabilites}
\end{equation}

where $\eta$ denotes the counting measure, i.e., the number of times a given tuple or sequence appears in the dataset $D$. 
If the counting measure in the denominator is zero, we will take the conditional density to be uniform. 
\end{comment}

%\textbf{Continuous Spaces.}

\begin{algorithm*}[t!]
\caption{Markov Balance-based Imitation Learning (MBIL)}
\label{alg:MBIL}
\textbf{Input}: Expert dataset of trajectories $D$ = $\{(s_{i} , a_{i} )\}_{i=1}^{n}$.\\
%\rishabh{Add a line about converting the trajectgories into (s, a), (s', a') pairs}
%\textbf{Parameter}: Optional list of parameters\\
\textbf{Output}: $\theta^{*}$.%, solution to \eqref{eqn:kernelRegularizedObjective}
\begin{algorithmic}[1] %enables line numbers
%\STATE Let $t=0$.
\STATE Initialize policy parameter $\theta$, expert MC parameter $\eta$, transition MDP parameter $\psi$.\\
\STATE Transform dataset $D$ into $(s, a, s', a')$ tuples, then store them in buffer $B$.
\STATE Train $P_{\eta}(s', a'|s, a)$ and $T_{\psi}(s'|s, a)$ by log-likelihood maximization, using the estimates provided by \eqref{eqn:FinalConditionalProbabilityEstimation}.
%\STATE Generate a tuple 
%\STATE Obtain $\hat{P}, \hat{T}$ in \eqref{eqn:kernelEstimators} via CKDE on $B$\\
\FOR{ $iter = 0, 1, \hdots$}
\STATE Sample a batch $b_{iter}$ of $(s, a, s', a')$ tuples from $B$.
\STATE Obtain $\hat{P_{\eta}}, \hat{T_{\psi}}$ on $b_{iter}$ using \eqref{eqn:FinalConditionalProbabilityEstimation} on $P_{\eta}$ and $T_{\psi}$ respectively.\\
\STATE Calculate empirical estimate of the objective function in \eqref{eqn:markovBalanceObjective} using all $(s, a, s', a')$ $\in b_{iter}$ as:
\begin{equation}
    %\sum_{(s, a) \in b_{iter}} \alpha [a - a_{\theta}]^2 +  \sum_{(s, a, s', a') \in b_{iter}} \beta \big[ log\hat{P_{\eta}}(s', a'|s, a) - log\pi_{\theta}(a'|s') - log\hat{T_{\psi}}(s'|s,a)\big]^{2}
    \sum_{(s, a, s', a') \in b_{iter}} \alpha \big[ \log\hat{P_{\eta}}(s', a'|s, a) - \log\pi_{\theta}(a'|s') - \log\hat{T_{\psi}} (s'|s,a)\big]^{2} +
    \sum_{(s, a) \in b_{iter}} \beta [\mathcal{L}\left(\pi_{D}(a \mid s), \pi_{\theta}(a \mid s)\right)]
    \label{eqn:empiricalObjectiveFunction}
\end{equation}
%, where $a_{\theta} \sim \pi_{\theta}(\cdot|s)$.
\STATE Update the policy parameter $\theta$ using gradient update to minimize the calculated empirical estimate of the objective function
\ENDFOR
\STATE \textbf{return} $\theta^{*}$
\end{algorithmic}
\end{algorithm*}

Estimating transition densities in continuous setting is challenging because each visited state would only appear once, and most states may never be visited in the dataset. This necessitates the use of more sophisticated conditional density estimation methods. These include non-parametric methods like Gaussian process conditional density estimation \citep{ckdeAlternative3} and Conditional Kernel Density Estimation (CKDE) \citep{NonparametricEconometrics}; semi-parametric methods like least squares conditional density estimation \citep{LSCKDE}; and parametric approaches such as mixture density networks (MDN) \citep{MDN} and normalizing flows \citep{dinh2017density}.

We use Normalizing Flows (NFs) because they outperform  density estimation techniques such as CKDE and MDN. Unlike CKDE and MDN, which can struggle with high-dimensional data and require careful tuning of bandwidths or mixture components, NFs leverage a series of invertible transformations to model complex distributions with high flexibility and precision. This  allows for exact likelihood computation, making NFs ideal for tasks requiring precise probability evaluation, as in our setup. Additionally, NFs can capture complex, multimodal distributions without predefined parametric forms, offering a more expressive framework \citep{papamakarios2021normalizing}. Furthermore, NFs with affine coupling flows are universal distribution approximators under suitable conditions \citep{pmlr-v235-draxler24a}.

Given a conditioning input $c \in C$, the conditional density estimation of $x \in \mathcal{X}$, $p(x|c)$ using a Conditional Normalizing Flow (CNF) is given by \citep{ardizzone2019guided}:
\begin{comment}
\begin{equation}
\resizebox{.99\hsize}{!}{$
    p_{X|C}(x|c)=p_{Z|C}(z|c)\left|\frac{\partial z}{\partial x}\right|
            =p_{Z|C}\left(g_\phi(x, c)\right)\left|\frac{\partial g_\phi(x, c)}{\partial x}\right|$}
\end{equation}
\begin{equation}
    p_{X|C}(x|c)=p_{Z|C}(z|c)\left|\frac{\partial z}{\partial x}\right|\\
            =p_{Z|C}\left(g_\phi(x, c)\right)\left|\frac{\partial g_\phi(x, c)}{\partial x}\right|
\end{equation}
\end{comment}
%p_{X|C}(x|c) & = p_{Z}(z|c)\left|\frac{\partial z}{\partial x}\right| \\
\begin{equation*}
%\begin{split}
p_{X|C}(x|c) = p_{Z}\left(g_{\phi, k}(x, c)\right)\left|\frac{\partial g_{\phi, k}(x, c)}{\partial x}\right|,
%\end{split}
\label{eqn:conditional_density_estimation_equation}
\end{equation*}
where $\left|\frac{\partial g_{\phi, k}(x, c)}{\partial x}\right|$ is the Jacobian determinant of $g_{\phi, k}$ evaluated at $x$ for the conditioning input $c$. Finally, the log-likelihood of $x$ given $c$ can be computed as:
\begin{equation}
\log p_{X|C}(x|c) = \log p_{Z}\left(g_{\phi, k}(x, c)\right) + \sum_{i=1}^{k} \log \left| \det \frac{\partial g_{i}}{\partial g_{i-1}} \right|
\label{eqn:FinalConditionalProbabilityEstimation}
\end{equation}

Figure \ref{fig:conditional_normalizing_flows} illustrates our conditional normalizing flow setup. The neural network $NN_{c}$ processes the conditioning input $c$, mapping it to a latent space variable $\tilde{c}$, which is then fed into each coupling block of our NF configuration. The specifics of the $i^{th}$  coupling block, for $i \in \{1, 2, \hdots, k\}$, are shown in Figure \ref{fig:coupling_block}. 
% Each $i^{th}$ 
Coupling block $i$ includes a neural network $NN_{i}$ that further transforms $\tilde{c}$ into $c'_{i}$ before passing it to the coupling flow.

\begin{figure*} % Use figure* for spanning both columns
    \centering
        \begin{tikzpicture}
        \hspace{0.2cm}
    \begin{customlegend}[legend columns=10]
    \addlegendimage{bcColor,mark=*, thick, mark options={solid,scale=1.5}}
    \addlegendentry{BC}
    \addlegendimage{vdiceColor,mark=triangle*,thick, mark options={solid,scale=1.5}}
    \addlegendentry{VDICE}
    \addlegendimage{rcalColor,mark=|, thick, mark options={solid,scale=1.5}}
    \addlegendentry{RCAL}
    \addlegendimage{edmColor,mark=asterisk, thick, mark options={solid,scale=1.5}}
    \addlegendentry{EDM}
    \addlegendimage{avrilColor,mark=x, thick, mark options={solid,scale=1.5}}
    \addlegendentry{AVRIL}
    \addlegendimage{dsfnColor,mark=triangle*, thick, mark options={solid,scale=1.5,rotate=180}}
    \addlegendentry{DSFN}
    \addlegendimage{ckilColor,mark=diamond*, thick, mark options={solid,scale=1.5}}
    \addlegendentry{MBIL}
    \addlegendimage{iqlearnColor,mark=pentagon*, thick, mark options={solid,scale=1.5}}
    \addlegendentry{IQ-Learn}
    \addlegendimage{expertColor,mark=|*, densely dashed, thick}
    \addlegendentry{Expert}
    \addlegendimage{randomColor,mark=|*, densely dashed, thick}
    \addlegendentry{Random}
    \end{customlegend}
    \end{tikzpicture}
    
    \begin{subfigure}[b]{0.31\textwidth}
        \includegraphics[width=\linewidth]
        %{Plots/Acrobot.png}
        {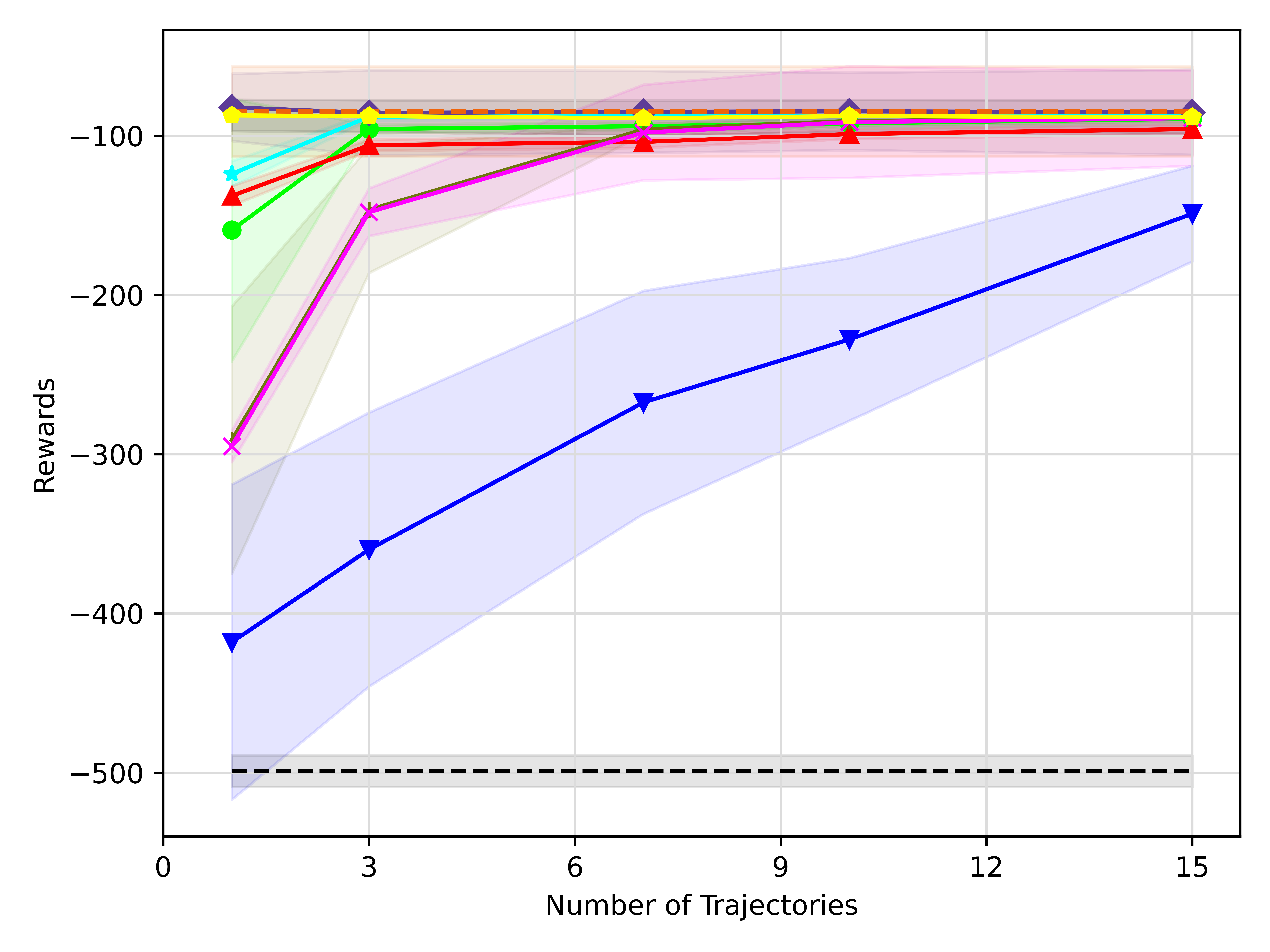}
        %{../Figures/AcrobotPlotWithoutLegend.png}%{AcrobotPlot.png} % Replace with your figure
        \caption{Acrobot}
    \end{subfigure}
    \hfill
    \begin{subfigure}[b]{0.31\textwidth}
        \includegraphics[width=\linewidth]
        %{Plots/Cartpole.png}
        {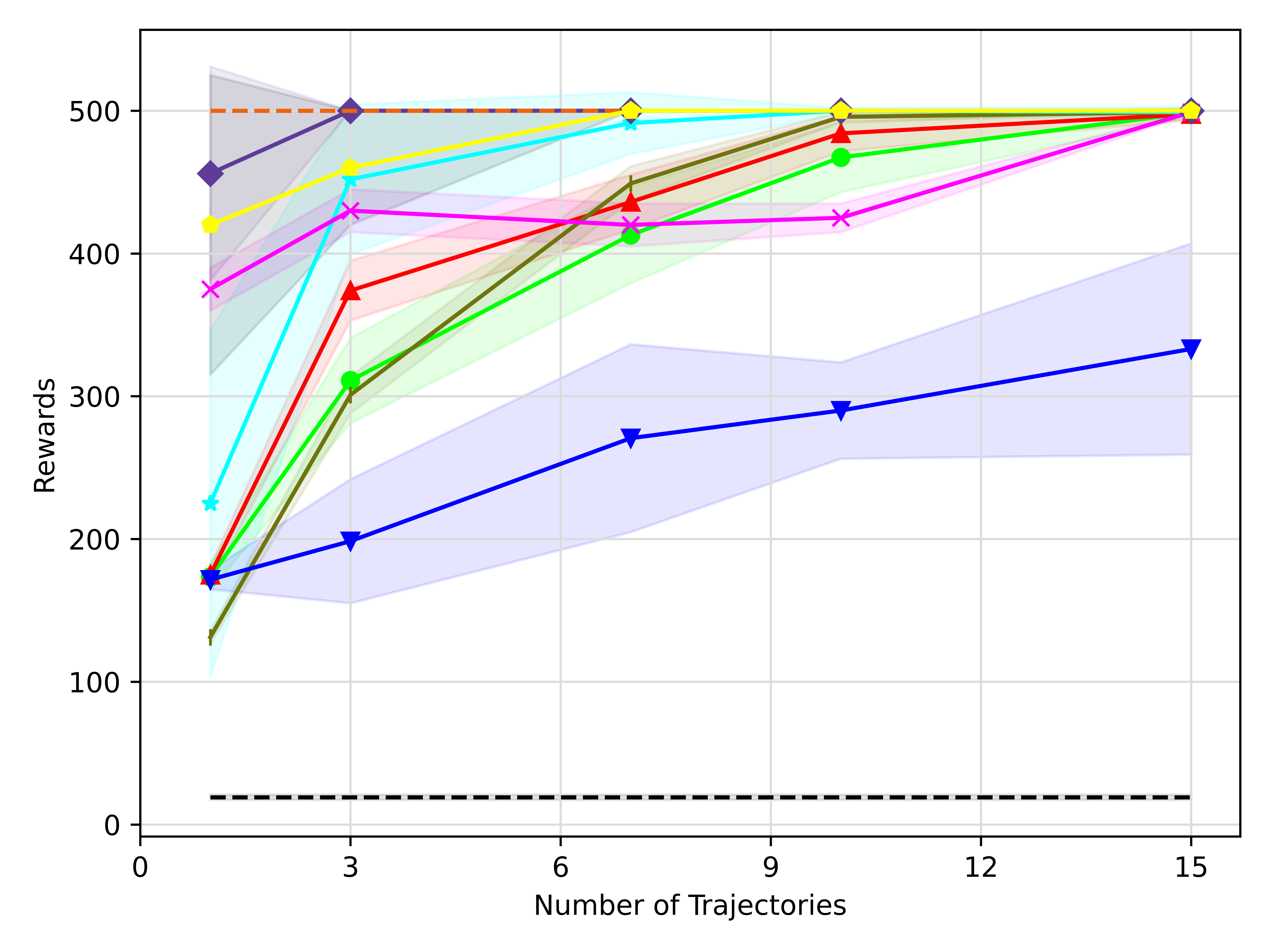}
        %{../Figures/CartPolePlotWithoutLegend.png}
        %{AnonymousSubmission/LaTeX/CartPolePlot.png} % Replace with your figure
        \caption{CartPole}
    \end{subfigure}
    \hfill
    \begin{subfigure}[b]{0.31\textwidth}
        \includegraphics[width=\linewidth]
        %{Plots/LunarLander.png}
        {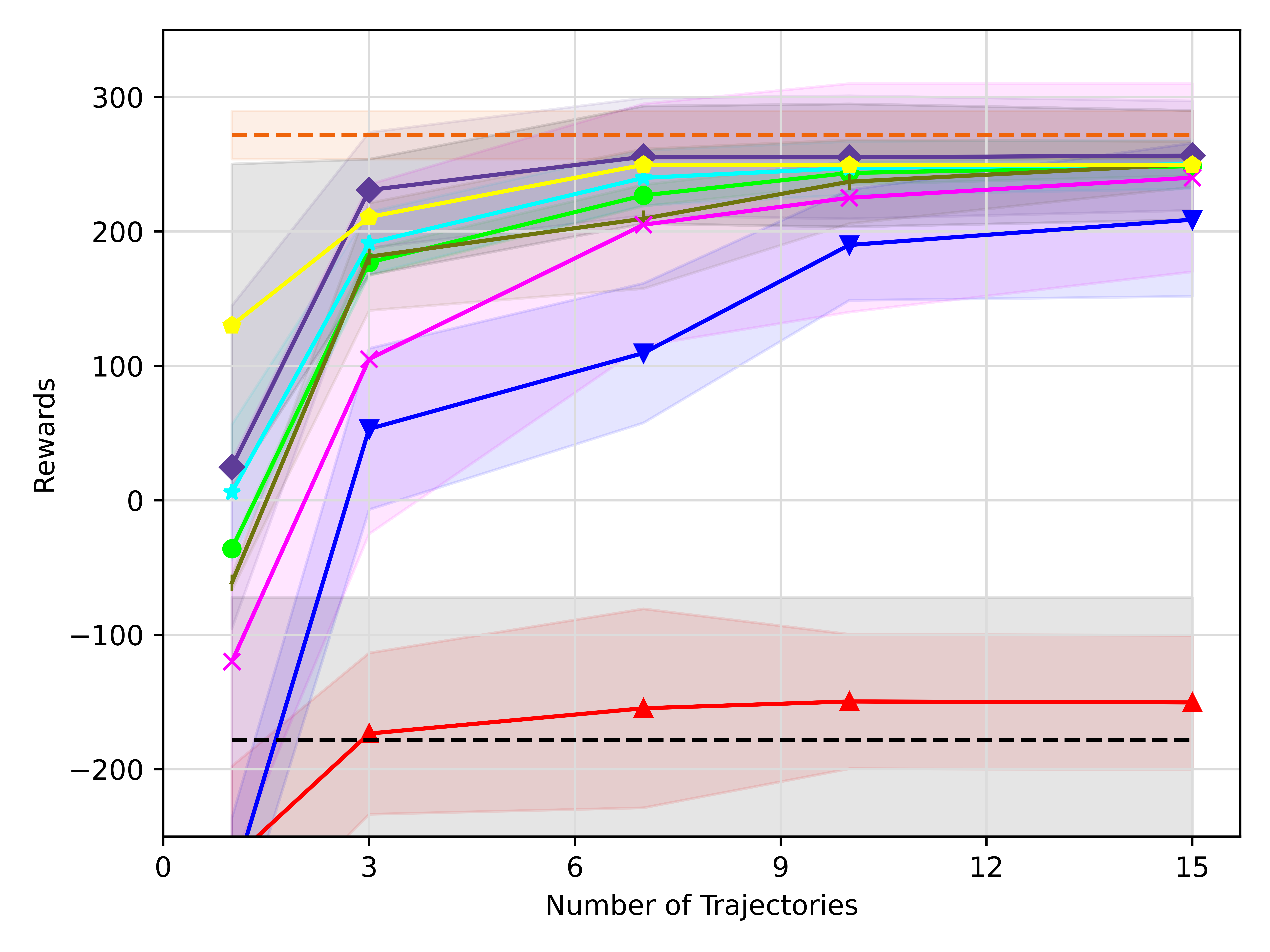}
        %{../Figures/LunarLanderPlotWithoutLegend.png}
        %{AnonymousSubmission/LaTeX/LunarLanderPlot.png} % Replace with your figure
        \caption{LunarLander}
    \end{subfigure}

    \caption{Average rewards achieved by benchmark IRL/IL/AIL and MBIL policies during real-time deployment plotted against the number of trajectories included in demonstration dataset $D$ (higher values indicate better performance).}
    \label{fig:controlplots}
\end{figure*}

\section{Experimental Results}\label{sec:results}
\textbf{Setup.} %We evaluate the empirical performance of the MBIL algorithm using the MuJoCo locomotion suite \citep{todorov2012mujoco}, available through OpenAI Gym \cite{openAIGym}. The diverse difficulty levels of these MuJoCo tasks make them a popular benchmark for imitation learning algorithms in continuous action domains. To construct the demonstration dataset $\emph{D}$, we source data from \citep{vdiceGithub}, where the Generative Adversarial Imitation Learning algorithm \citep{gail} was utilized to generate the data.
We evaluate the empirical performance of our MBIL algorithm using the MuJoCo locomotion suite \citep{todorov2012mujoco} and the classic control suite from OpenAI Gym \cite{openAIGym}. MuJoCo tasks, with their varying difficulty levels, are a popular benchmark for IL in continuous action domains, while classic control environments like LunarLander are used to assess IL algorithms in discrete action domains. For constructing the demonstration dataset $\emph{D}$, we use data from \citep{vdiceGithub}, where the Generative Adversarial Imitation Learning algorithm \citep{gail} was applied for MuJoCo tasks. For classic control tasks, we utilize pre-trained and hyperparameter-optimized agents from the RL Baselines Zoo \citep{rl-zoo3}, employing a PPO agent for LunarLander-v2, a DQN agent for CartPole-v1, and an A2C agent for Acrobot-v1.

%\textbf{Baseline Algorithms.} We compare the performance of MBIL (Algorithm \ref{alg:MBIL}) against several offline IL/IRL/AIL baselines, including recent state-of-the-art approaches such as Behavioral Cloning (BC), ValueDICE \citep{VDICE_iclr}, SoftDICE \citep{sun2021softdice}, and ODICE \citep{mao2024odice}. \textcolor{red}{We also include IQ-Learn \citep{garg2021iq}, a prominent model-free offline IRL algorithm, in our comparison.} It is important to note that some recent offline IRL/IL methods, such as CLARE \citep{clare}, either require additional diverse data or allow for interactions with the environment within their off-policy frameworks. Since our approach strictly utilizes only expert data without additional interactions or supplementary data, we exclude these methods to ensure a fair comparison. A more detailed discussion is provided in Appendix \ref{app}.

\textbf{Implementation.} %In \eqref{eqn:empiricalObjectiveFunction}, the policy $\pi_{\theta}$ is modeled as a Gaussian distribution with parameters determined by a neural network (NN). This NN consists of two hidden layers, each using the Rectified Linear Unit (ReLU) activation function. The output layer has twice the number of nodes as the action dimension, representing each action dimension's mean and standard deviation. 
A neural network (NN) with two hidden layers, each using the ReLU activation function, is used for representing the policy. For tasks with discrete actions (e.g., Classic Control tasks), the output layer has a dimension equal to the number of action dimensions and uses a softmax function to generate a probability distribution over actions given a state. 
%In contrast, for MuJoCo tasks involving continuous actions, the output layer has twice the number of nodes as the action dimension, providing the means and standard deviations for each action dimension. The policy is then modeled as a Gaussian distribution with these parameters produced by the NN.
In contrast, for MuJoCo tasks with continuous actions, the output consists of two separate layers: one for the mean and another for the standard deviations, each with a size equal to the action dimension. The policy is then modeled as a Gaussian distribution, with these parameters generated by the neural network.
Training is performed with the Adam optimizer \citep{adam}. Detailed implementation, including hyperparameters for MBIL and benchmark algorithms, are provided in the Appendix.

The transition dynamics models $P_{\eta}$ and $T_{\psi}$ are implemented using RealNVPs \citep{dinh2017density}. We employ the publicly available framework version 0.2 \citep{freia}, utilizing their GLOWCouplingBlocks implementation. Detailed parameters for these models are provided in Appendix. A central aspect of our approach is performing imitation learning in the low data regime. To facilitate this, we introduce Gaussian noise as a regularizer for training the expert MC and transition MDP, which enhances training stability with limited data.

\textbf{Results.}
When given sufficient demonstration data, all benchmarks can achieve performance comparable to optimized demonstration agents. Therefore, we test the algorithms' ability to handle limited data. 
Several recent offline IRL/IL methods, such as CLARE \citep{clare}, DemoDICE \citep{kim2022demodice}, and MILO \citep{chang2021mitigating}, depend on additional diverse data, while others, like OPOLO \citep{zhu2020off}, allow for interactions with the environment within their off-policy frameworks. Given that our setting strictly considers only offline expert data without access to additional interactions or supplementary datasets, we exclude methods that rely on such resources from our comparisons to ensure a fair evaluation.

\textbf{Classic Control Tasks.} Inspired by \citep{SBIL}, we trained algorithms until convergence on datasets of $1$, $3$, $7$, $10$, or $15$ trajectories sampled from a pool of $1000$ expert trajectories and recorded the average scores over $300$ episodes for each algorithm, repeating this process $10$ times with varied initializations and trajectories. We compare our MBIL algorithm (Algorithm \ref{alg:MBIL}) against various offline IRL/IL/AIL baselines, including BC, ValueDICE (VDICE), RCAL, EDM, AVRIL, DSFN, and the state-of-the-art model-free offline IRL algorithm IQ-Learn. We use log-likelihood maximization based BC as $\mathcal{L}(\cdot)$ in \eqref{eqn:empiricalObjectiveFunction} in our Classic Control experiments. 

Figure \ref{fig:controlplots} illustrates the average rewards obtained by each algorithm as the demonstration dataset size increases in the Acrobot, CartPole, and LunarLander environments. The results highlight MBIL's ability to learn effective policies, consistently outperforming the baseline algorithms, particularly when data is limited. MBIL achieves near-expert-level performance in these environments with at most three trajectories and, remarkably, can reach near-expert performance in the CartPole environment with just a single trajectory. Notably, MBIL generally outperforms all baseline algorithms across these environments, with IQ-Learn showing performance closest to MBIL. Additionally, off-policy adaptations of online algorithms, such as VDICE and DSFN, do not maintain the same level of consistent performance as their inherently online counterparts. This highlights the need for more than just adopting online algorithms in offline scenarios. Moreover, the difficulty in estimating the expectation of an exponential distribution may contribute to VDICE's relative underperformance compared to MBIL and other methods.

\begin{figure*} % Use figure* for spanning both columns
    \centering
        \begin{tikzpicture}
    \begin{customlegend}[legend columns=-1]
    \addlegendimage{bcColor,mark=, thick, mark options={solid,scale=1.5}}
    \addlegendentry{BC}
    \addlegendimage{vdiceColor,mark=,thick, mark options={solid,scale=1.5}}
    \addlegendentry{VDICE}
    \addlegendimage{softdiceColor,mark=, thick, mark options={solid,scale=1.5}}
    \addlegendentry{SoftDICE}
    \addlegendimage{odiceColor,mark=, thick, mark options={solid,scale=1.5}}
    \addlegendentry{ODICE}
    \addlegendimage{ckilColor,mark=, thick, mark options={solid,scale=1.5}}
    \addlegendentry{MBIL}
    \addlegendimage{expertColor,mark=|*, densely dashed, thick}
    \addlegendentry{Expert}
    \end{customlegend}
    \end{tikzpicture}
    
    \begin{subfigure}[b]{0.245\textwidth}
        \includegraphics[width=\linewidth]
        %{Plots/Hopper-v2.png}
        {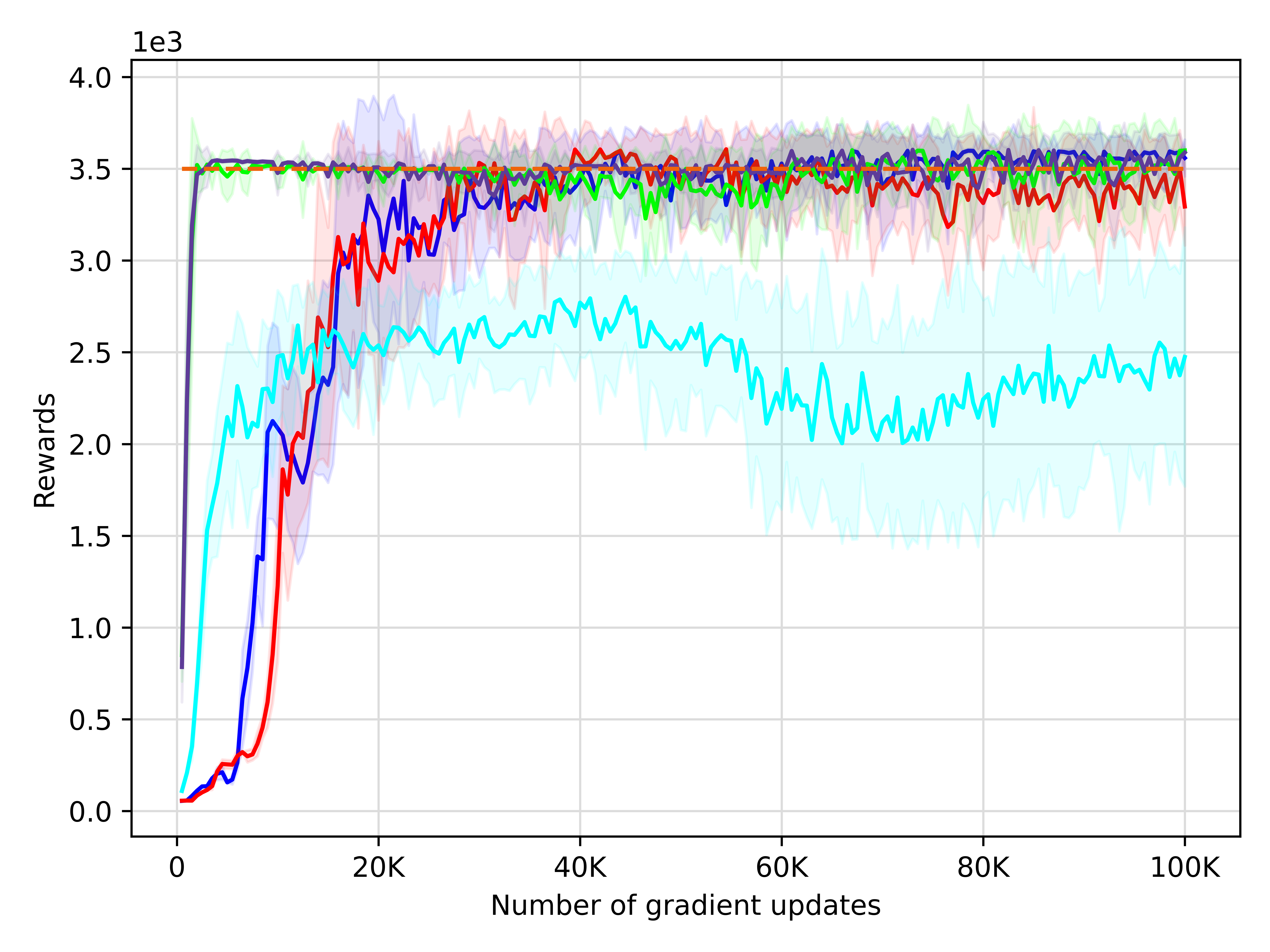}
        %{AnonymousSubmission/LaTeX/AcrobotPlot.png} % Replace with your figure
        \caption{Hopper}
    \end{subfigure}
    \hfill
    \begin{subfigure}[b]{0.245\textwidth}
        \includegraphics[width=\linewidth]
        %{Plots/Ant-v2.png}
        {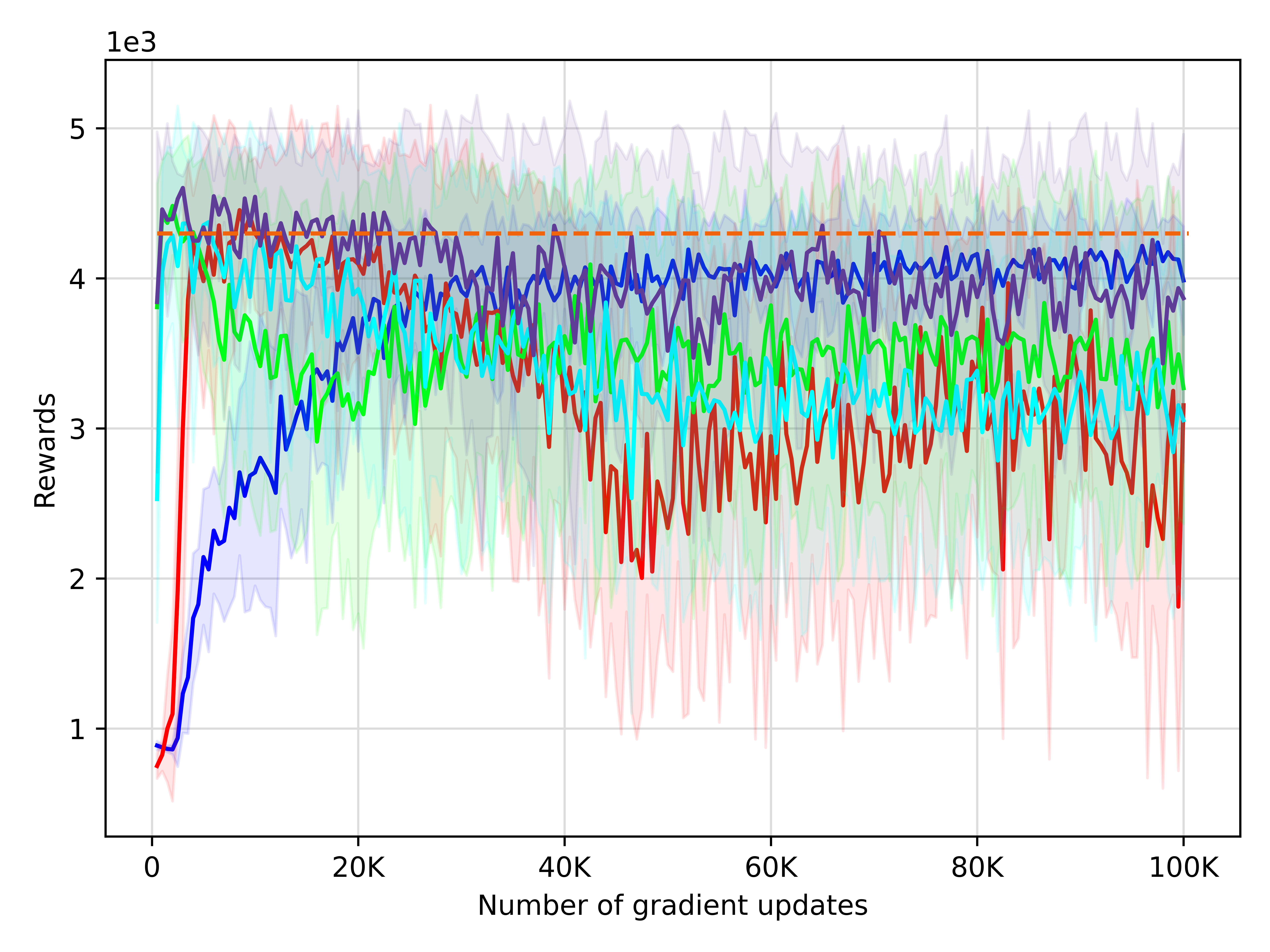}
        %{AnonymousSubmission/LaTeX/CartPolePlot.png} % Replace with your figure
        \caption{Ant}
    \end{subfigure}
    \hfill
    \begin{subfigure}[b]{0.245\textwidth}
        \includegraphics[width=\linewidth]
        %{Plots/HalfCheetah-v2.png}
        {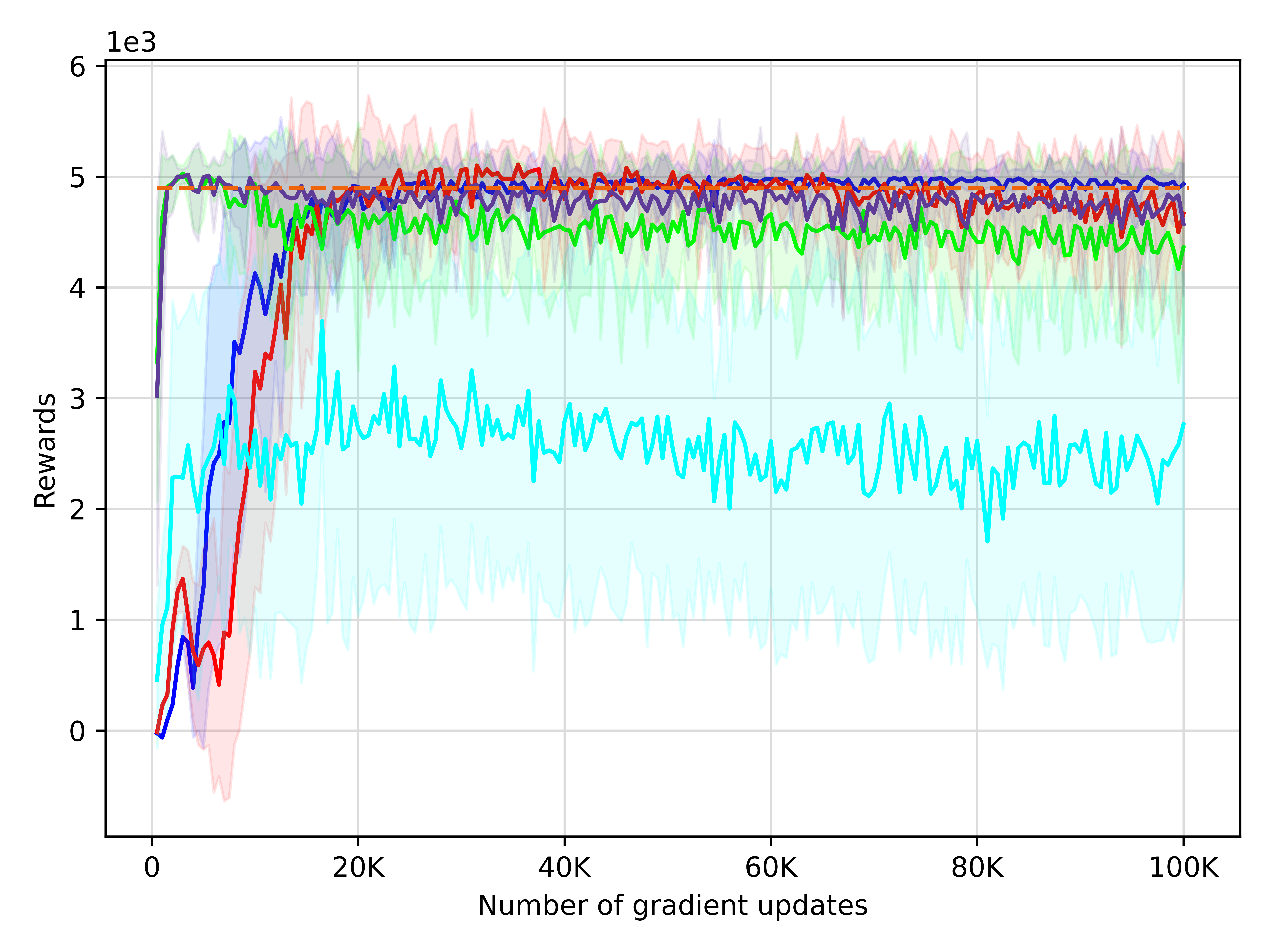}
        %{AnonymousSubmission/LaTeX/LunarLanderPlot.png} % Replace with your figure
        \caption{HalfCheetah}
    \end{subfigure}
        \begin{subfigure}[b]{0.245\textwidth}
        \includegraphics[width=\linewidth]
        %{Plots/Walker2d-v2.png}
        {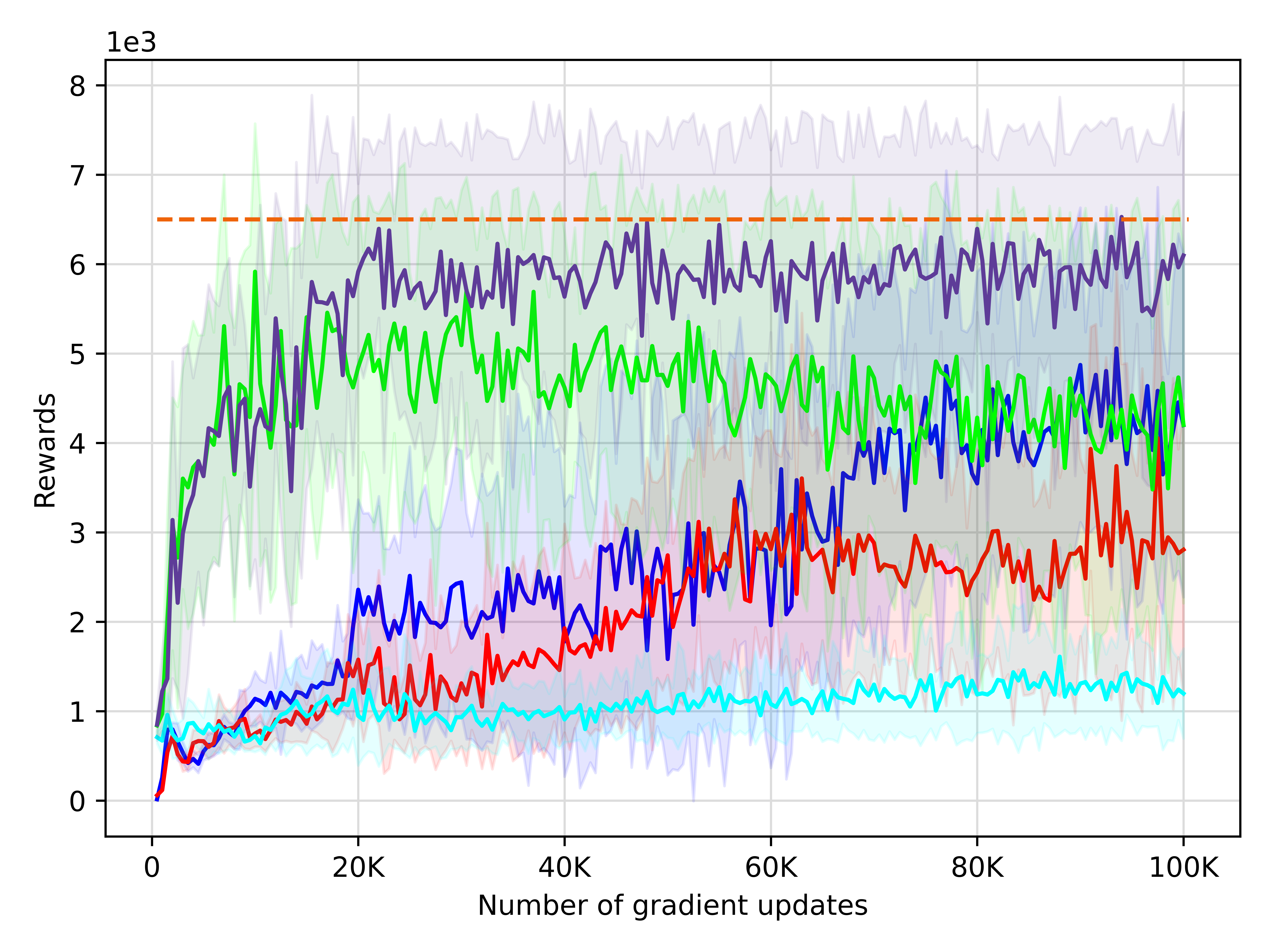}
        %{AnonymousSubmission/LaTeX/LunarLanderPlot.png} % Replace with your figure
        \caption{Walker2d}
    \end{subfigure}
    \caption{Average rewards achieved by benchmark and MBIL policies against the number of gradient updates on MuJoCo tasks with $1$ expert trajectory (higher values indicate better performance).}
    \label{fig:mujocoplots}
\end{figure*}

\begin{figure}[t]
\centering
\includegraphics[width=0.9\columnwidth]
%{Plots/Walker2d-v2_ablation.png}
{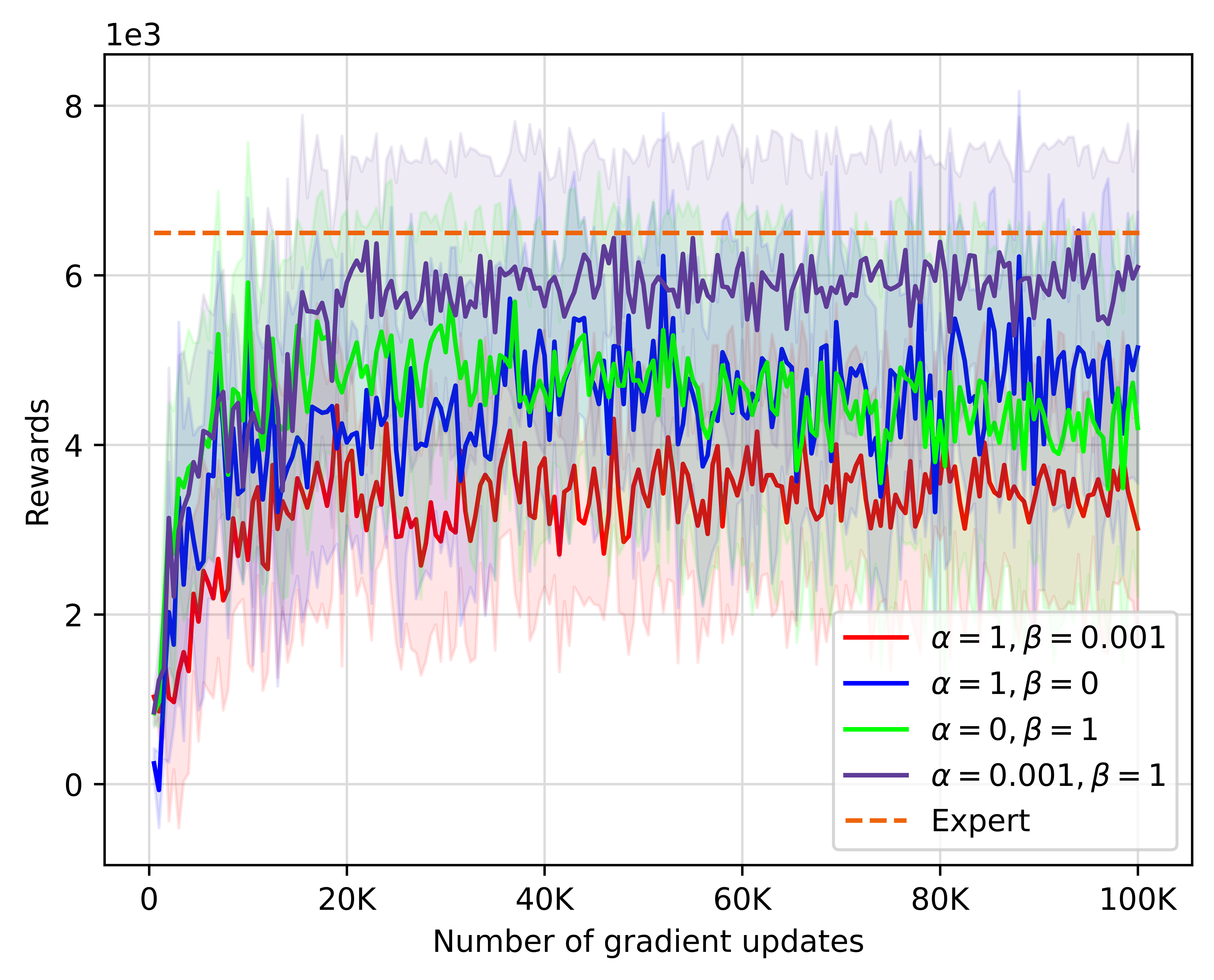}
% Reduce the figure size so that it is slightly narrower than the column. Don't use precise values for figure width.This setup will avoid overfull boxes.
\caption{Ablation study: Average rewards achieved by policies trained using different combinations of $\alpha$ and $\beta$ values in the MBIL objective against the number of gradient updates on MuJoCo tasks with $1$ expert trajectory on Walker2d environment (higher values indicate better performance).}
\label{fig:ablation}
\end{figure}

\textbf{MuJoCo Tasks.} Following the methodology outlined in \citep{VDICE_iclr, sun2021softdice}, we use a single demonstration trajectory, validate performance every $500$ training iterations across $10$ episodes, and report means and standard deviations from $5$ random seeds. We compare MBIL (Algorithm \ref{alg:MBIL}) against strong baselines for locomotion tasks, including BC, ValueDICE, SoftDICE, and ODICE. IQ-Learn is excluded due to its poor performance in high-dimensional MuJoCo tasks with continuous action spaces in the strictly offline setting, as it suffers from exploding Q-functions \citep{al-hafez2023lsiq}. We also exclude EDM and AVRIL because they are incompatible with continuous action spaces. 
Notably, while baselines like SoftDICE and ValueDICE, which use log-likelihood maximization for BC and show its poor performance, MSE-based BC, as demonstrated by \citep{li2022rethinking}, performs well in MuJoCo tasks and is therefore used for BC in our comparison. Furthermore, we employ this mean-squared error (MSE) BC loss
%, defined as $\left[a - a_{\theta}\right]^2$, where $a_{\theta} \sim \pi_{\theta}(\cdot|s)$ for $s, a \sim \rho_{\pi_{D}}$, and use this 
as $\mathcal{L}(\cdot)$ in \eqref{eqn:empiricalObjectiveFunction} in our MuJoCo experiments.

Figure \ref{fig:mujocoplots} compares the performance of MBIL with strong baselines on MuJoCo tasks in the offline IL literature. MBIL consistently achieves near-expert performance on all tasks with just a single trajectory in the dataset. Among the baselines, BC and SoftDICE show strong performance. However, BC's performance deteriorates in the Ant and Walker2d environments as the number of gradient updates increases, despite using orthogonal regularization \citep{brock2018large}, a trend also noted in \citep{li2022rethinking}. In contrast, MBIL maintains robust performance throughout the learning process by integrating dynamics loss with the policy loss. SoftDICE encounters difficulties in the Walker2d environment, where absorbing states are more common than Ant and HalfCheetah. Its assumption of a value of $0$ for these states can introduce reward bias, hindering performance since the values of these states also need to be learned \citep{al-hafez2023lsiq}. The performance decline of ValueDICE is likely due to biased gradient updates in its learning process \citep{sun2021softdice}. ODICE attempts to address the issue of conflicting gradients in DICE methods by incorporating orthogonal gradient updates, but this approach may not have been effective for these tasks. Overall, the results highlight our framework's effectiveness, which combines policy loss and dynamics loss to facilitate effective learning.

\textbf{Ablation study. } 
In our framework, the objective function comprises two components: 1) the policy loss and 2) the dynamics loss. We investigate how each component affects the performance of our algorithm on the Walker2d task by varying the $\alpha$ and $\beta$ coefficients in \eqref{eqn:markovBalanceObjective}, with results shown in Figure \ref{fig:ablation}. Specifically, we evaluate the following configurations: $\alpha=1, \beta=0$ (dynamics loss only), $\alpha=0, \beta=1$ (policy loss only, representing BC in this experiment), $\alpha=1, \beta=0.001$ (higher relative weight on dynamics loss), and $\alpha=0.001, \beta=1$ (higher relative weight on policy loss). Our findings reveal that while both policy loss alone and dynamics loss alone perform similarly, with dynamics loss performing slightly better as gradient updates progress, it is the combination of policy loss and dynamics loss, with a higher weight assigned to the policy loss, that enables us to achieve near-expert performance using only one expert trajectory.

\section{Conclusion}\label{sec:conclusion}

In this paper, we address the strictly batch imitation learning problem in a continuous state and action space setting, i.e., we do not assume access to any further interactions or supplementary data. We present a Markov balance-based imitation learning algorithm that combines it with supervised learning-based behavior cloning while using conditional normalizing flows for density estimation. Our method does not rely on reward model estimation
%nor assumes that the expert's policy is optimal in any way. 
and avoids stationary distribution matching which suffer from termination and reward bias. From numerical experiments, we see that our proposed algorithm does as well or much better than any state-of-the-art algorithm across a variety of Classic Control and MuJoCo environments. In fact, as can be seen in  further results presented in the Appendix, it handles the distribution shift issue more effectively than other strictly batch imitation learning algorithms. Future research could benefit from deriving both asymptotic and non-asymptotic sample complexity bounds, which are scarce in current literature, and from exploring extensions to handle suboptimal data cases.

%[In this paper, we present a framework for Imitation Learning in a strictly offline setting. Unlike IRL methods, our algorithm avoids reward learning, and unlike DICE methods, it does not rely on matching the stationary distribution of state-action pairs as stationarity fails to hold in practice, thus avoiding the termination and reward bias associated with this approach. It also eliminates the alternating optimization, as done in DICE methods. Instead, our framework uses conditional normalizing flow estimators for the transition densities of the MDP and the induced Markov chain, optimizing the policy at every transition to adhere to the Markov balance equation. Our algorithm achieves state-of-the-art performance compared to offline IL, IRL, and AIL methods and does not require additional datasets or environmental interactions. Additionally, it is worth noting that any imitation learning algorithm operating with access to a very limited training dataset encounters the challenge of distribution shift in a strictly offline setup \citep{ajay2023is}. However, our experimental findings suggest that our algorithm handles the distribution shift issue more effectively than other strictly batch imitation learning algorithms. Please refer to Appendix for further insights. Future research could benefit from deriving both asymptotic and non-asymptotic sample complexity bounds, which are scarce in current literature, and from exploring extensions to handle suboptimal data cases.]

%\newpage
%\bibliography{main}  % Comment or remove this line

\begin{thebibliography}{59}
\providecommand{\natexlab}[1]{#1}

\bibitem[{Abbeel and Ng(2004)}]{AndrewNG1}
Abbeel, P.; and Ng, A.~Y. 2004.
\newblock Apprenticeship Learning via Inverse Reinforcement Learning.
\newblock In \emph{Proceedings of the Twenty-First International Conference on Machine Learning}.

\bibitem[{Al-Hafez et~al.(2023)Al-Hafez, Tateo, Arenz, Zhao, and Peters}]{al-hafez2023lsiq}
Al-Hafez, F.; Tateo, D.; Arenz, O.; Zhao, G.; and Peters, J. 2023.
\newblock {LS}-{IQ}: Implicit Reward Regularization for Inverse Reinforcement Learning.
\newblock In \emph{The Eleventh International Conference on Learning Representations}.

\bibitem[{Ardizzone et~al.(2018-2022)Ardizzone, Bungert, Draxler, Köthe, Kruse, Schmier, and Sorrenson}]{freia}
Ardizzone, L.; Bungert, T.; Draxler, F.; Köthe, U.; Kruse, J.; Schmier, R.; and Sorrenson, P. 2018-2022.
\newblock {Framework for Easily Invertible Architectures (FrEIA)}.

\bibitem[{Ardizzone et~al.(2019)Ardizzone, L{\"u}th, Kruse, Rother, and K{\"o}the}]{ardizzone2019guided}
Ardizzone, L.; L{\"u}th, C.; Kruse, J.; Rother, C.; and K{\"o}the, U. 2019.
\newblock Guided image generation with conditional invertible neural networks.
\newblock \emph{arXiv preprint arXiv:1907.02392}.

\bibitem[{Arora and Doshi(2021)}]{aroraAndDoshi2021}
Arora, S.; and Doshi, P. 2021.
\newblock A survey of inverse reinforcement learning: Challenges, methods and progress.
\newblock \emph{Artificial Intelligence}, 297: 103500.

\bibitem[{Baheri(2023)}]{baheri2023understanding}
Baheri, A. 2023.
\newblock Understanding Reward Ambiguity Through Optimal Transport Theory in Inverse Reinforcement Learning.
\newblock In \emph{NeurIPS 2023 Workshop Optimal Transport and Machine Learning}.

\bibitem[{Barnes et~al.(2024)Barnes, Abueg, Lange, Deeds, Trader, Molitor, Wulfmeier, and O'Banion}]{barnes2024massively}
Barnes, M.; Abueg, M.; Lange, O.~F.; Deeds, M.; Trader, J.; Molitor, D.; Wulfmeier, M.; and O'Banion, S. 2024.
\newblock Massively Scalable Inverse Reinforcement Learning in Google Maps.
\newblock In \emph{The Twelfth International Conference on Learning Representations}.

\bibitem[{Bishop(1994)}]{MDN}
Bishop, C. 1994.
\newblock Mixture density networks.
\newblock Technical report, Aston University.

\bibitem[{Brock, Donahue, and Simonyan(2018)}]{brock2018large}
Brock, A.; Donahue, J.; and Simonyan, K. 2018.
\newblock Large scale GAN training for high fidelity natural image synthesis.
\newblock \emph{arXiv preprint arXiv:1809.11096}.

\bibitem[{Brockman et~al.(2016)Brockman, Cheung, Pettersson, Schneider, Schulman, Tang, and Zaremba}]{openAIGym}
Brockman, G.; Cheung, V.; Pettersson, L.; Schneider, J.; Schulman, J.; Tang, J.; and Zaremba, W. 2016.
\newblock Openai gym.
\newblock \emph{arXiv preprint arXiv:1606.01540}.

\bibitem[{Chan and van~der Schaar(2021)}]{chan2021}
Chan, A.~J.; and van~der Schaar, M. 2021.
\newblock Scalable Bayesian Inverse Reinforcement Learning.
\newblock In \emph{International Conference on Learning Representations}.

\bibitem[{Chang et~al.(2021)Chang, Uehara, Sreenivas, Kidambi, and Sun}]{chang2021mitigating}
Chang, J.~D.; Uehara, M.; Sreenivas, D.; Kidambi, R.; and Sun, W. 2021.
\newblock Mitigating covariate shift in imitation learning via offline data without great coverage.
\newblock \emph{arXiv preprint arXiv:2106.03207}.

\bibitem[{Dinh, Krueger, and Bengio(2014)}]{dinh2014nice}
Dinh, L.; Krueger, D.; and Bengio, Y. 2014.
\newblock Nice: Non-linear independent components estimation.
\newblock \emph{arXiv preprint arXiv:1410.8516}.

\bibitem[{Dinh, Sohl-Dickstein, and Bengio(2017)}]{dinh2017density}
Dinh, L.; Sohl-Dickstein, J.; and Bengio, S. 2017.
\newblock Density estimation using Real {NVP}.
\newblock In \emph{International Conference on Learning Representations}.

\bibitem[{Draxler et~al.(2024)Draxler, Wahl, Schnoerr, and Koethe}]{pmlr-v235-draxler24a}
Draxler, F.; Wahl, S.; Schnoerr, C.; and Koethe, U. 2024.
\newblock On the Universality of Volume-Preserving and Coupling-Based Normalizing Flows.
\newblock In Salakhutdinov, R.; Kolter, Z.; Heller, K.; Weller, A.; Oliver, N.; Scarlett, J.; and Berkenkamp, F., eds., \emph{Proceedings of the 41st International Conference on Machine Learning}, volume 235 of \emph{Proceedings of Machine Learning Research}, 11613--11641. PMLR.

\bibitem[{Dutordoir et~al.(2018)Dutordoir, Salimbeni, Hensman, and Deisenroth}]{ckdeAlternative3}
Dutordoir, V.; Salimbeni, H.; Hensman, J.; and Deisenroth, M. 2018.
\newblock Gaussian process conditional density estimation.
\newblock \emph{Advances in neural information processing systems}, 31.

\bibitem[{Fu, Luo, and Levine(2018)}]{fu2018AIL}
Fu, J.; Luo, K.; and Levine, S. 2018.
\newblock Learning Robust Rewards with Adverserial Inverse Reinforcement Learning.
\newblock In \emph{International Conference on Learning Representations}.

\bibitem[{Garg et~al.(2021)Garg, Chakraborty, Cundy, Song, and Ermon}]{garg2021iq}
Garg, D.; Chakraborty, S.; Cundy, C.; Song, J.; and Ermon, S. 2021.
\newblock Iq-learn: Inverse soft-q learning for imitation.
\newblock \emph{Advances in Neural Information Processing Systems}, 34: 4028--4039.

\bibitem[{Ho and Ermon(2016)}]{gail}
Ho, J.; and Ermon, S. 2016.
\newblock Generative adversarial imitation learning.
\newblock \emph{Advances in neural information processing systems}, 29.

\bibitem[{Jarrett, Bica, and van~der Schaar(2020)}]{SBIL}
Jarrett, D.; Bica, I.; and van~der Schaar, M. 2020.
\newblock Strictly batch imitation learning by energy-based distribution matching.
\newblock \emph{Advances in Neural Information Processing Systems}, 33: 7354--7365.

\bibitem[{Jumper et~al.(2021)Jumper, Evans, Pritzel, Green, Figurnov, Ronneberger, Tunyasuvunakool, Bates, {\v{Z}}{\'\i}dek, Potapenko et~al.}]{jumper2021highly}
Jumper, J.; Evans, R.; Pritzel, A.; Green, T.; Figurnov, M.; Ronneberger, O.; Tunyasuvunakool, K.; Bates, R.; {\v{Z}}{\'\i}dek, A.; Potapenko, A.; et~al. 2021.
\newblock Highly accurate protein structure prediction with AlphaFold.
\newblock \emph{Nature}, 596(7873): 583--589.

\bibitem[{Ke et~al.(2021)Ke, Choudhury, Barnes, Sun, Lee, and Srinivasa}]{fDivAIL}
Ke, L.; Choudhury, S.; Barnes, M.; Sun, W.; Lee, G.; and Srinivasa, S. 2021.
\newblock Imitation learning as f-divergence minimization.
\newblock In \emph{Algorithmic Foundations of Robotics XIV: Proceedings of the Fourteenth Workshop on the Algorithmic Foundations of Robotics 14}, 313--329. Springer.

\bibitem[{Kim et~al.(2022)Kim, Seo, Lee, Jeon, Hwang, Yang, and Kim}]{kim2022demodice}
Kim, G.-H.; Seo, S.; Lee, J.; Jeon, W.; Hwang, H.; Yang, H.; and Kim, K.-E. 2022.
\newblock Demodice: Offline imitation learning with supplementary imperfect demonstrations.
\newblock In \emph{International Conference on Learning Representations}.

\bibitem[{Kingma and Ba(2015)}]{adam}
Kingma, D.~P.; and Ba, J. 2015.
\newblock Adam: {A} Method for Stochastic Optimization.
\newblock In \emph{3rd International Conference on Learning Representations}.

\bibitem[{Kingma and Dhariwal(2018)}]{kingma2018glow}
Kingma, D.~P.; and Dhariwal, P. 2018.
\newblock Glow: Generative flow with invertible 1x1 convolutions.
\newblock \emph{Advances in neural information processing systems}, 31.

\bibitem[{Klein, Geist, and Pietquin(2011)}]{klein2011}
Klein, E.; Geist, M.; and Pietquin, O. 2011.
\newblock Batch, off-policy and model-free apprenticeship learning.
\newblock In \emph{European Workshop on Reinforcement Learning}, 285--296. Springer.

\bibitem[{Klein et~al.(2012)Klein, Geist, Piot, and Pietquin}]{klein2012}
Klein, E.; Geist, M.; Piot, B.; and Pietquin, O. 2012.
\newblock Inverse reinforcement learning through structured classification.
\newblock \emph{Advances in neural information processing systems}, 25.

\bibitem[{Kostrikov et~al.(2018)Kostrikov, Agrawal, Dwibedi, Levine, and Tompson}]{kostrikov2018discriminator}
Kostrikov, I.; Agrawal, K.~K.; Dwibedi, D.; Levine, S.; and Tompson, J. 2018.
\newblock Discriminator-actor-critic: Addressing sample inefficiency and reward bias in adversarial imitation learning.
\newblock \emph{arXiv preprint arXiv:1809.02925}.

\bibitem[{Kostrikov, Nachum, and Tompson(2020{\natexlab{a}})}]{VDICE_iclr}
Kostrikov, I.; Nachum, O.; and Tompson, J. 2020{\natexlab{a}}.
\newblock Imitation Learning via Off-Policy Distribution Matching.
\newblock In \emph{International Conference on Learning Representations}.

\bibitem[{Kostrikov, Nachum, and Tompson(2020{\natexlab{b}})}]{vdiceGithub}
Kostrikov, I.; Nachum, O.; and Tompson, J. 2020{\natexlab{b}}.
\newblock Imitation Learning via Off-Policy Distribution Matching.
\newblock \url{https://github.com/google-research/google-research/tree/master/value_dice}.

\bibitem[{Lee, Srinivasan, and Doshi-Velez(2019)}]{lee2019}
Lee, D.; Srinivasan, S.; and Doshi-Velez, F. 2019.
\newblock Truly Batch Apprenticeship Learning with Deep Successor Features.
\newblock In \emph{Proceedings of the Twenty-Eighth International Joint Conference on Artificial Intelligence}, 5909--5915.

\bibitem[{Li and Racine(2006)}]{NonparametricEconometrics}
Li, Q.; and Racine, J.~S. 2006.
\newblock \emph{{Nonparametric Econometrics: Theory and Practice}}.
\newblock Number 8355 in Economics Books. Princeton University Press.

\bibitem[{Li et~al.(2022)Li, Xu, Yu, and Luo}]{li2022rethinking}
Li, Z.; Xu, T.; Yu, Y.; and Luo, Z.-Q. 2022.
\newblock Rethinking ValueDice: Does it really improve performance?
\newblock \emph{arXiv preprint arXiv:2202.02468}.

\bibitem[{Ma et~al.(2022)Ma, Shen, Jayaraman, and Bastani}]{ma2022versatile}
Ma, Y.; Shen, A.; Jayaraman, D.; and Bastani, O. 2022.
\newblock Versatile offline imitation from observations and examples via regularized state-occupancy matching.
\newblock In \emph{International Conference on Machine Learning}, 14639--14663. PMLR.

\bibitem[{Mao et~al.(2024)Mao, Xu, Zhang, and Zhan}]{mao2024odice}
Mao, L.; Xu, H.; Zhang, W.; and Zhan, X. 2024.
\newblock Odice: Revealing the mystery of distribution correction estimation via orthogonal-gradient update.
\newblock \emph{arXiv preprint arXiv:2402.00348}.

\bibitem[{Mnih et~al.(2015)Mnih, Kavukcuoglu, Silver, Rusu, Veness, Bellemare, Graves, Riedmiller, Fidjeland, Ostrovski et~al.}]{mnih2015human}
Mnih, V.; Kavukcuoglu, K.; Silver, D.; Rusu, A.~A.; Veness, J.; Bellemare, M.~G.; Graves, A.; Riedmiller, M.; Fidjeland, A.~K.; Ostrovski, G.; et~al. 2015.
\newblock Human-level control through deep reinforcement learning.
\newblock \emph{nature}, 518(7540): 529--533.

\bibitem[{Ng and Russell(2000)}]{AndrewNG2}
Ng, A.~Y.; and Russell, S.~J. 2000.
\newblock Algorithms for Inverse Reinforcement Learning.
\newblock In \emph{Proceedings of the Seventeenth International Conference on Machine Learning}, 663–670.

\bibitem[{Papamakarios et~al.(2021)Papamakarios, Nalisnick, Rezende, Mohamed, and Lakshminarayanan}]{papamakarios2021normalizing}
Papamakarios, G.; Nalisnick, E.; Rezende, D.~J.; Mohamed, S.; and Lakshminarayanan, B. 2021.
\newblock Normalizing flows for probabilistic modeling and inference.
\newblock \emph{Journal of Machine Learning Research}, 22(57): 1--64.

\bibitem[{Papamakarios, Pavlakou, and Murray(2017)}]{papamakarios2017masked}
Papamakarios, G.; Pavlakou, T.; and Murray, I. 2017.
\newblock Masked autoregressive flow for density estimation.
\newblock \emph{Advances in neural information processing systems}, 30.

\bibitem[{Piot, Geist, and Pietquin(2014)}]{bcLimitation2}
Piot, B.; Geist, M.; and Pietquin, O. 2014.
\newblock Boosted and Reward-Regularized Classification for Apprenticeship Learning.
\newblock In \emph{Proceedings of the 2014 International Conference on Autonomous Agents and Multi-Agent Systems}, 1249–1256.

\bibitem[{Piot, Geist, and Pietquin(2016)}]{BCMod2}
Piot, B.; Geist, M.; and Pietquin, O. 2016.
\newblock Bridging the gap between imitation learning and inverse reinforcement learning.
\newblock \emph{IEEE transactions on neural networks and learning systems}, 28(8): 1814--1826.

\bibitem[{Pomerleau(1988)}]{firstBC}
Pomerleau, D.~A. 1988.
\newblock Alvinn: An autonomous land vehicle in a neural network.
\newblock \emph{Advances in neural information processing systems}, 1: 305–313.

\bibitem[{Raffin(2020)}]{rl-zoo3}
Raffin, A. 2020.
\newblock RL Baselines3 Zoo.
\newblock \url{https://github.com/DLR-RM/rl-baselines3-zoo}.

\bibitem[{Ross and Bagnell(2010)}]{bcLimitation1}
Ross, S.; and Bagnell, D. 2010.
\newblock Efficient reductions for imitation learning.
\newblock In \emph{Proceedings of the thirteenth international conference on artificial intelligence and statistics}, 661--668. JMLR Workshop and Conference Proceedings.

\bibitem[{Ross, Gordon, and Bagnell(2011)}]{BCMod1}
Ross, S.; Gordon, G.; and Bagnell, D. 2011.
\newblock A Reduction of Imitation Learning and Structured Prediction to No-Regret Online Learning.
\newblock In \emph{Proceedings of the Fourteenth International Conference on Artificial Intelligence and Statistics}, volume~15, 627--635.

\bibitem[{Silver et~al.(2016)Silver, Huang, Maddison, Guez, Sifre, Van Den~Driessche, Schrittwieser, Antonoglou, Panneershelvam, Lanctot et~al.}]{silver2016mastering}
Silver, D.; Huang, A.; Maddison, C.~J.; Guez, A.; Sifre, L.; Van Den~Driessche, G.; Schrittwieser, J.; Antonoglou, I.; Panneershelvam, V.; Lanctot, M.; et~al. 2016.
\newblock Mastering the game of Go with deep neural networks and tree search.
\newblock \emph{nature}, 529(7587): 484--489.

\bibitem[{Silver et~al.(2021)Silver, Singh, Precup, and Sutton}]{silver2021reward}
Silver, D.; Singh, S.; Precup, D.; and Sutton, R.~S. 2021.
\newblock Reward is enough.
\newblock \emph{Artificial Intelligence}, 299: 103535.

\bibitem[{Sugiyama et~al.(2010)Sugiyama, Takeuchi, Suzuki, Kanamori, Hachiya, and Okanohara}]{LSCKDE}
Sugiyama, M.; Takeuchi, I.; Suzuki, T.; Kanamori, T.; Hachiya, H.; and Okanohara, D. 2010.
\newblock Least-squares conditional density estimation.
\newblock \emph{IEICE Transactions on Information and Systems}, 93(3): 583--594.

\bibitem[{Sun et~al.(2021)Sun, Mahajan, Hofmann, and Whiteson}]{sun2021softdice}
Sun, M.; Mahajan, A.; Hofmann, K.; and Whiteson, S. 2021.
\newblock Softdice for imitation learning: Rethinking off-policy distribution matching.
\newblock \emph{arXiv preprint arXiv:2106.03155}.

\bibitem[{Swamy et~al.(2021)Swamy, Choudhury, Bagnell, and Wu}]{sbilCritique}
Swamy, G.; Choudhury, S.; Bagnell, J.~A.; and Wu, Z.~S. 2021.
\newblock A Critique of Strictly Batch Imitation Learning.
\newblock \emph{arXiv preprint arXiv:2110.02063}.

\bibitem[{Todorov, Erez, and Tassa(2012)}]{todorov2012mujoco}
Todorov, E.; Erez, T.; and Tassa, Y. 2012.
\newblock Mujoco: A physics engine for model-based control.
\newblock In \emph{2012 IEEE/RSJ international conference on intelligent robots and systems}, 5026--5033. IEEE.

\bibitem[{Vinyals et~al.(2019)Vinyals, Babuschkin, Czarnecki, Mathieu, Dudzik, Chung, Choi, Powell, Ewalds, Georgiev et~al.}]{vinyals2019grandmaster}
Vinyals, O.; Babuschkin, I.; Czarnecki, W.~M.; Mathieu, M.; Dudzik, A.; Chung, J.; Choi, D.~H.; Powell, R.; Ewalds, T.; Georgiev, P.; et~al. 2019.
\newblock Grandmaster level in StarCraft II using multi-agent reinforcement learning.
\newblock \emph{Nature}, 575(7782): 350--354.

\bibitem[{Xu et~al.(2022)Xu, Zhan, Yin, and Qin}]{xu2022discriminator}
Xu, H.; Zhan, X.; Yin, H.; and Qin, H. 2022.
\newblock Discriminator-weighted offline imitation learning from suboptimal demonstrations.
\newblock In \emph{International Conference on Machine Learning}, 24725--24742. PMLR.

\bibitem[{Yue et~al.(2023{\natexlab{a}})Yue, Wang, Shao, Zhang, Lin, Ren, and Zhang}]{yue2023clare}
Yue, S.; Wang, G.; Shao, W.; Zhang, Z.; Lin, S.; Ren, J.; and Zhang, J. 2023{\natexlab{a}}.
\newblock {CLARE}: Conservative Model-Based Reward Learning for Offline Inverse Reinforcement Learning.
\newblock In \emph{The Eleventh International Conference on Learning Representations}.

\bibitem[{Yue et~al.(2023{\natexlab{b}})Yue, Wang, Shao, Zhang, Lin, Ren, and Zhang}]{clare}
Yue, S.; Wang, G.; Shao, W.; Zhang, Z.; Lin, S.; Ren, J.; and Zhang, J. 2023{\natexlab{b}}.
\newblock {CLARE}: Conservative Model-Based Reward Learning for Offline Inverse Reinforcement Learning.
\newblock In \emph{The Eleventh International Conference on Learning Representations}.

\bibitem[{Yue and Le(2018)}]{ImitationLearningPresentation}
Yue, Y.; and Le, H.~M. 2018.
\newblock Imitation learning (Tutorial).
\newblock \emph{International Conference on Machine Learning (ICML)}.

\bibitem[{Zeng et~al.(2023)Zeng, Li, Garcia, and Hong}]{Siliang}
Zeng, S.; Li, C.; Garcia, A.; and Hong, M. 2023.
\newblock Understanding Expertise through Demonstrations: A Maximum Likelihood Framework for Offline Inverse Reinforcement Learning.
\newblock \emph{arXiv preprint arXiv:2302.07457}.

\bibitem[{Zhu et~al.(2020)Zhu, Lin, Dai, and Zhou}]{zhu2020off}
Zhu, Z.; Lin, K.; Dai, B.; and Zhou, J. 2020.
\newblock Off-policy imitation learning from observations.
\newblock \emph{Advances in neural information processing systems}, 33: 12402--12413.

\bibitem[{Ziebart et~al.(2008)Ziebart, Maas, Bagnell, and Dey}]{maxEntRL}
Ziebart, B.~D.; Maas, A.; Bagnell, J.~A.; and Dey, A.~K. 2008.
\newblock Maximum Entropy Inverse Reinforcement Learning.
\newblock In \emph{Proceedings of the 23rd National Conference on Artificial Intelligence - Volume 3}, 1433–1438. AAAI Press.

\end{thebibliography}

       % Add this line

\end{document}